%% file: spice_mixer.tex
\documentclass[sigconf,nonacm]{acmart}

\usepackage[ruled, vlined, linesnumbered]{algorithm2e}
\usepackage{amsmath}
\usepackage{array}
\usepackage{booktabs}
\usepackage{float}
\usepackage{fvextra}
\usepackage{graphicx}
\usepackage{makecell}
\usepackage{multirow}
\usepackage{paralist}
\usepackage{stfloats}
\usepackage{siunitx}
\usepackage{subcaption}
\usepackage[most]{tcolorbox}
\usepackage{textcomp}
\usepackage{xcolor}
\def\BibTeX{{\rm B\kern-.05em{\sc i\kern-.025em b}\kern-.08em
    T\kern-.1667em\lower.7ex\hbox{E}\kern-.125emX}}

\newcommand{\squeezeSpace}{\vspace{-0.22cm}}
\newcommand{\squeezeSection}{\vspace*{-0.22cm}}
\newcommand{\squeezeSubSection}{\vspace*{-0.12cm}}
\newcommand{\squeezeSubSubSection}{\vspace*{-0.12cm}}

\newboolean{anonymous}
\setboolean{anonymous}{false} % Set to true for anonymous, false otherwise

\newboolean{addappendix}
\setboolean{addappendix}{false} % Set to true to show appendix with netlists

\newcolumntype{C}[1]{>{\centering\arraybackslash}p{#1}}

\newtcolorbox{prompt}[1][]{
    enhanced,
    drop shadow={black!50!white},
    coltitle=black,
    top=4pt,
    bottom=-1pt,
    attach boxed title to top left={xshift=1.5em,yshift=-\tcboxedtitleheight/2},
    boxed title style={size=small, colback=lightgray},
    fonttitle=\bfseries\scriptsize,
    title={#1}
}

\newcommand{\spm}{\emph{SPICE\-Mixer}}
\definecolor{darkgreen}{rgb}{0.0, 0.5, 0.0}
\definecolor{inputcolor}{rgb}{0.0, 0.4, 0.8}
\definecolor{outputcolor}{rgb}{0.0, 0.6, 0.4}
\definecolor{internalcolor}{rgb}{0.6, 0.4, 0.8}
\definecolor{supplycolor}{rgb}{0.8, 0.4, 0.0}

\definecolor{darkblue}{RGB}{0, 0, 139}
\newcommand{\str}[1]{\textcolor{darkblue}{`#1'}}

\begin{document}

% remove indentation of itemize environments
\setdefaultleftmargin{1em}{0pt}{}{}{}{}

\title{\spm{} -- Netlist-Level Circuit Evolution}

\ifthenelse{\boolean{anonymous}}{
  \author{Anonymous Authors}
  \affiliation{%
    \institution{Anonymous Institution(s)}
    \country{}}
  \renewcommand{\shortauthors}{Anonymous}
  \setcopyright{acmlicensed}
  \copyrightyear{2026}
  \acmYear{2026}
  \acmDOI{XXXXXXX.XXXXXXX}
  \acmConference[DAC '26]{Design Automation Conference 2026}{July 26-29, 2026}{Long Beach, CA}
  \widowpenalty=100     % allow a widow (last line on next page)
  \clubpenalty=100      % allow a club line (first line on next page)
}{
    \author{Stefan Uhlich\textsuperscript{1,2}, Andrea Bonetti\textsuperscript{1}, Arun Venkitaraman\textsuperscript{1}, Chia-Yu Hsieh\textsuperscript{1},\newline Yağız Gençer\textsuperscript{2,3}, Mustafa Emre Gürsoy\textsuperscript{2,3}, Ryoga Matsuo\textsuperscript{1,3}, Lorenzo Servadei\textsuperscript{1,4}}
    \affiliation{%
    \institution{\textsuperscript{1}\textit{SonyAI, Switzerland} \quad
    \textsuperscript{2}\textit{Sony Semiconductor Solutions Europe, Germany} \quad
    \textsuperscript{3}\textit{EPFL, Switzerland} \quad
    \textsuperscript{4}\textit{TU Munich, Germany}}}
    \renewcommand{\shortauthors}{Uhlich, Bonetti, Venkitaraman, Hsieh, Gençer, Gürsoy, Matsuo, Servadei}
}

\begin{abstract}
We present \spm{}, a genetic algorithm that synthesizes circuits by directly evolving SPICE netlists. \spm{} operates on individual netlist lines, making it compatible with arbitrary components and subcircuits and enabling general-purpose genetic operators: crossover, mutation, and pruning, all applied directly at the netlist level. To support these operators, we normalize each netlist by enforcing consistent net naming (inputs, outputs, supplies, and internal nets) and by sorting components and nets into a fixed order, so that similar circuit structures appear at similar line positions. This normalized netlist format improves the effectiveness of crossover, mutation, and pruning. We demonstrate \spm{} by synthesizing standard cells (e.g., NAND2 and latch) and by designing OpAmps that meet specified targets. Across tasks, \spm{} matches or exceeds recent synthesis methods while requiring substantially fewer simulations.
\end{abstract}




\maketitle

%%%%%%%%%%%%%%%%%%%%%%%%%%%%%%%%%%%%%%%%%%%%%%%%%%%%%%%%%%
\squeezeSection\section{Introduction}
%%%%%%%%%%%%%%%%%%%%%%%%%%%%%%%%%%%%%%%%%%%%%%%%%%%%%%%%%%

\begin{figure*}
    \centering
    \includegraphics[width=0.9\linewidth,trim=0 5 0 20]{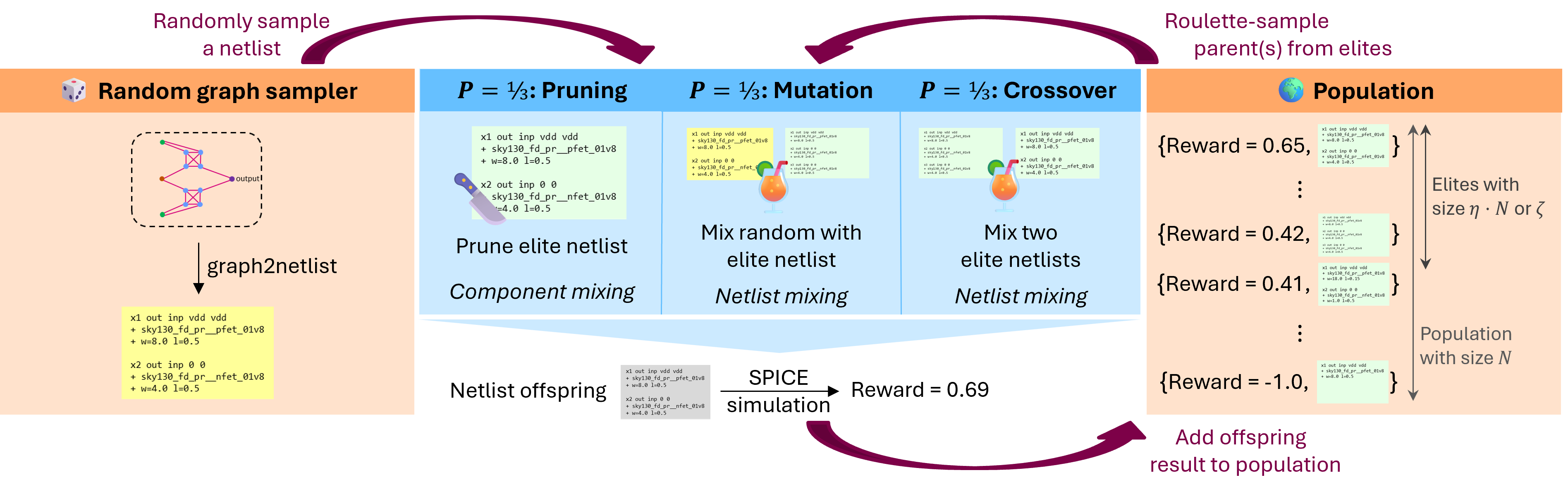}
    \caption{Proposed \spm{} approach. With equal probability, one of three genetic operators (pruning, mutation, or crossover) applies \emph{component} or \emph{netlist mixing} to generate an offspring netlist, which is then evaluated with a SPICE simulation.}
    \Description{Proposed \spm{} approach. With equal probability, one of three genetic operators (pruning, mutation, or crossover) applies \emph{component} or \emph{netlist mixing} to generate an offspring netlist, which is then evaluated with a SPICE simulation.}
    \label{fig:spicemixer_approach}
    \squeezeSpace\squeezeSpace
\end{figure*}

Despite recent progress, circuit design remains a challenging and time-intensive process, often relying heavily on expert knowledge and manual tuning~\cite{razavi2017design,jespers2017systematic}. While machine learning and optimization techniques have advanced automated digital design, progress in analog automation has been slower due to the complexity of analog design spaces.

In this work, we present \spm{}, a \emph{genetic algorithm} (GA) that evolves SPICE netlists. The general approach is illustrated in Fig.~\ref{fig:spicemixer_approach}. Unlike prior methods~\cite{koza1996automated,trefzer2006evolution,campilho2024analog}, \spm{} operates directly on SPICE netlists, which motivates its name. By applying genetic operations such as crossover, mutation, and pruning at the netlist level, \spm{} avoids the need for abstract graph representations or custom GA chromosome encodings of circuits. This straightforward yet robust approach enables efficient exploration of the circuit design space and can be applied to any circuit for which a testbench defines a reward. In our experiments, it successfully synthesizes standard cells and operational amplifiers, outperforming prior methods.
 The main contributions of this paper are:
\begin{compactitem}
    \item A netlist-level GA for analog synthesis, \spm{}, whose main hyperparameters (elite set size, mutation probability, and pruning rate) are tuned to balance crossover, mutation, and pruning.
    \item Results on synthesizing standard digital cells (inverter, latch, AND2, NAND2, OR2, NOR2) and comparison against published baselines from~\cite{uhlich2024graco}.
    \item Results on designing operational amplifiers (OpAmps) assembled from common subcircuits (current mirrors and differential pairs) that meet target specifications.
\end{compactitem}
This paper is structured as follows: Sec.~\ref{sec:related_work} reviews related work and positions \spm{} within that context, including a brief overview of the GraCo framework from~\cite{uhlich2024graco}, which underpins our approach. Sec.~\ref{sec:spm} details the \spm{} method and its genetic netlist operations. Sec.~\ref{sec:results} presents experimental results and compares \spm{} to two GraCo-based methods and the Covariance Matrix Adaptation Evolution Strategy (CMA-ES)~\cite{hansen2016cma}. Finally, Sec.~\ref{sec:conclusions} gives our conclusions and an outlook on future work.

%%%%%%%%%%%%%%%%%%%%%%%%%%%%%%%%%%%%%%%%%%%%%%%%%%%%%%%%%%
\squeezeSection\section{Related Work}
\label{sec:related_work}
%%%%%%%%%%%%%%%%%%%%%%%%%%%%%%%%%%%%%%%%%%%%%%%%%%%%%%%%%%

%%%%%%%%%%%%%%%%%%%%%%%%%%%%%%%%%%%%%%%%%%%%%%%%%%%%%%%%%%
\subsection{Circuit Synthesis Methods}
\label{sec:related_work:subsec:circuit_synthesis}
%%%%%%%%%%%%%%%%%%%%%%%%%%%%%%%%%%%%%%%%%%%%%%%%%%%%%%%%%%
Circuit synthesis generally involves two main steps: (i) determining the appropriate topology, meaning the selection of circuit components and their connections, and (ii) selecting the optimal parameter sizes for these components~\cite{zadeh2025generative}.

Substantial progress has been made in component sizing, as demonstrated in~\cite{yang2017smart,settaluri2020autockt,budak2021dnn,budak2023apostle,lyu2024study,momeni2025locality}. A recent comprehensive overview can be found in~\cite{lyu2024study}. However, identifying the correct topology remains a challenging and largely open problem. In recent years, several promising approaches have been proposed. For example, autoregressive models like AnalogCoder~\cite{lai2025analogcoder} or AnalogXpert~\cite{zhang2024analogxpert} can translate task descriptions and specifications directly into PySpice netlists or select subcircuit blocks and their connections, while AnalogGenie~\cite{gao2025analoggenie} models Eulerian cycles autoregressively to generate circuit topologies. The analogy between circuits and graphs has also been explored in depth, including the use of graph neural networks trained via supervised or reinforcement learning, as seen in~\cite{zhao2022deep,dong2023cktgnn,uhlich2024graco}. Other strategies leverage predefined library components, as demonstrated in~\cite{meissner2014feats,zhao2020automated}.

Another notable branch of work applies GAs to circuit synthesis, the category into which our proposed \spm{} method also falls. GA has been successfully applied not only for sizing~\cite{noren2001analog,yengui2012hybrid,barari2014analog,kwon2023circuit,rashid2024machine} but also for topology search~\cite{ning1991seas,kruiskamp1995darwin,kool2019buy,trefzer2006evolution,das2007gapsys,torres2010robust,sapargaliyev2010challenging,campilho2024analog}. While some of these works focus on specific applications, such as digital filter design~\cite{das2007gapsys,torres2010robust,sapargaliyev2010challenging} or operational amplifiers~\cite{kruiskamp1995darwin}, they generally rely on specialized representations (most often chromosome encodings~\cite{campilho2024analog}, but sometimes also computer programs~\cite{koza1996automated} or connection matrices~\cite{trefzer2006evolution}) to represent the circuit topology. In contrast, \spm{} operates directly on the netlist, which serves as its genome representation. This avoids the need for a specific chromosome encoding to support new components and makes the method naturally applicable to any component type or process development kit (PDK). It also handles circuits of varying sizes without requiring a variable-length chromosome representation as in~\cite{campilho2024analog}. Such variable-length encodings make genetic operations more involved and often require specialized, context-aware crossover algorithms.

%%%%%%%%%%%%%%%%%%%%%%%%%%%%%%%%%%%%%%%%%%%%%%%%%%%%%%%%%%
\squeezeSubSection\subsection{GraCo Framework}
\label{sec:related_work:subsec:graco}
%%%%%%%%%%%%%%%%%%%%%%%%%%%%%%%%%%%%%%%%%%%%%%%%%%%%%%%%%%

Since \spm{} builds upon GraCo~\cite{uhlich2024graco}, we begin with a brief overview of this framework before describing our approach.

GraCo is a framework for the automated synthesis of integrated circuits (ICs). It constructs circuit topologies by representing them as graphs, which are then translated into netlists and evaluated using SPICE simulations. To guide the sampling process, GraCo applies design constraints and consistency checks.

In its reinforcement learning setup, GraCo uses a reward function that maps SPICE simulation outcomes to a scalar between $-1$ and $1$. The reward is the average of multiple components, each comparing a simulated metric (such as voltage, timing, or power) to its target via a quadratic error. A saturation function ensures that each component reaches $1$ only when the specification is fully met. Once all components reach $1$, the circuit is considered valid and sampling stops. The same reward is used in \spm{}, also known as fitness function in the context of genetic algorithms.

GraCo supports two reinforcement learning methods: \emph{REINFORCE with Leave-One-Out} (RLOO)~\cite{williams1992simple,kool2019buy} and \emph{Evolution Strategies} (ES)~\cite{salimans2017evolution}. As reported in~\cite{uhlich2024graco}, ES has shown greater effectiveness in exploring complex design spaces and achieving superior synthesis outcomes as it can avoid collapsing too early to a suboptimal solution. Additionally, a random graph sampler was used in~\cite{uhlich2024graco} as a baseline. This sampler generates circuits by uniformly selecting components, connections, and sizing parameters, without relying on feedback or prior knowledge. In \spm{}, this random sampler is employed to create the initial netlist population, which is then refined through crossover, mutation, or pruning. Notably, the random sampler also applies explicit wiring rules and consistency checks, making it a good initial method to generate the starting population despite its simplicity.

%%%%%%%%%%%%%%%%%%%%%%%%%%%%%%%%%%%%%%%%%%%%%%%%%%%%%%%%%%
\squeezeSection\section{\spm{} Approach}
\label{sec:spm}
%%%%%%%%%%%%%%%%%%%%%%%%%%%%%%%%%%%%%%%%%%%%%%%%%%%%%%%%%%

We now describe \spm{}, which applies a genetic algorithm to synthesize circuits. We first introduce the normalized netlist format used, which makes the two core genetic operations, \emph{netlist mixing} and \emph{component mixing}, both reasonable and effective. Finally, we present the complete approach, as summarized in Fig.~\ref{fig:spicemixer_approach}.

%%%%%%%%%%%%%%%%%%%%%%%%%%%%%%%%%%%%%%%%%%%%%%%%%%%%%%%%%%
\squeezeSubSection\subsection{Normalization of Netlists}
\label{sec:spm:subsec:normalization}
%%%%%%%%%%%%%%%%%%%%%%%%%%%%%%%%%%%%%%%%%%%%%%%%%%%%%%%%%%

\input{example_netlist_mixing}

\input{example_netlist_pruning}

All netlists follow a standardized format designed to enhance the performance of the genetic operations. Specifically, we apply consistent net naming conventions: \texttt{net\_input\_\%d}, \texttt{net\_supply\_\%d}, \texttt{net\_output\_\%d}, and \texttt{net\_internal\_\%d} to represent input, supply, output, and internal nets, respectively. In addition, we apply the following normalizations, which uniformize the netlist but preserve the underlying circuit:
\begin{compactitem}
    \item \emph{Line sorting}: organizes netlist lines (i.e., component definitions) into \emph{input}, \emph{internal}, and \emph{output} blocks based on the net names they connect to, and sorts lines within each block alphabetically.
    \item \emph{Net sorting}: sorts drain and source net names alphabetically for NMOS/PMOS transistors, exploiting the drain/source symmetry in the device model.
    \item \emph{Internal net renumbering}: ensures that internal nets are numbered sequentially starting from zero in the order of first appearance.
    \item \emph{Component renumbering}: ensures that component indices are numbered sequentially starting from zero.
\end{compactitem}
These steps produce a structured, normalized netlist that enables more effective processing. Analogous to a DNA sequence~\cite{alberts14}, components serving similar functions (e.g., connecting input nets to internal nets) tend to appear in consistent positions across netlists, which improves the effectiveness of the genetic operations. In addition, this normalization ensures that circuits with the same topology and device sizes map to the same netlist representation, so the elite set does not store redundant variants of the same solution, which increases its diversity. In Sec.~\ref{sec:results:subsec:ablation}, we show that normalization has a substantial impact on performance.

%%%%%%%%%%%%%%%%%%%%%%%%%%%%%%%%%%%%%%%%%%%%%%%%%%%%%%%%%%
\squeezeSubSection\subsection{Genetic Operators}
\label{sec:spm:subsec:genetic_operators}
%%%%%%%%%%%%%%%%%%%%%%%%%%%%%%%%%%%%%%%%%%%%%%%%%%%%%%%%%%

\subsubsection{Crossover -- Mixing of two Elite Netlists}
\spm{} maintains an elite set of high-reward circuits and selects two of them as parents for crossover. To create a new offspring netlist, we apply \emph{netlist mixing}, choosing at each line index one of the following actions with equal probability:
\begin{compactitem}
    \item Add a line from the first netlist (if available).
    \item Add a line from the second netlist (if available).
    \item Add lines from both netlists (if available).
    \item Skip both lines.
\end{compactitem}
The offspring is then normalized as described in Sec.~\ref{sec:spm:subsec:normalization}. An example of the full crossover operation is shown in Fig.~\ref{fig:example_netlist_mixing}.

Two key points are worth noting: First, this approach roughly maintains the netlist length since all actions are equally probable. However, netlists can still shrink or grow over time if doing so improves the observed reward as we sample from the elite set. Second, merging is effective because we use normalized netlists, as discussed above. Normalization aligns the netlist structure of circuits from the elite set or from the random sampler, which makes \emph{netlist mixing} more meaningful.

\squeezeSubSubSection\subsubsection{Mutation -- Mixing of an Elite and a Random Netlist}
Mutation also uses \emph{netlist mixing} but pairs an elite parent with a randomly generated netlist, which can introduce new components, wirings, and sizings.

Importantly, in the mutation operator we bias \emph{netlist mixing} toward the elite parent. We treat the second (right) input netlist as the elite one and increase the probability of the action ``add a line from the second netlist''. Mixing elite and random netlists with equal probability would often produce weaker offsprings that do not enter the elite set. We control the bias with a mutation probability $\alpha$, which is the expected fraction of lines that do not come from the elite netlist (i.e., they are taken from the random netlist or skipped). In our experiments, we set $\alpha = 0.3$, so that, on average, 70\% of the lines are copied from the elite parent and 30\% are modified. This value of $\alpha$ was determined empirically by analyzing which genetic operations most often produced elite circuits. A detailed discussion is given in Sec.~\ref{sec:results:subsec:hyperparameters}.

\squeezeSubSubSection\subsubsection{Pruning -- Mixing of two Components}
To produce a new, more compact netlist, we apply \emph{component pruning}. This involves mixing two component definitions, that is, two netlist lines, which have the same number of elements (i.e., components with the same total count of nets and parameters), and retaining only the newly generated line, replacing the two original ones. Specifically, for each element, we randomly choose with equal probability whether to keep it from the first or the second component definition. An example is shown in Fig.~\ref{fig:example_component_mixing}.

Overall, we apply this pruning step up to $1 + \lfloor \beta \cdot L \rfloor$ times, where $L$ is the number of lines in the netlist, $\beta$ is a pruning rate, and $\lfloor . \rfloor$ denotes rounding down to the nearest integer. For example, applying pruning to a netlist with $10$ components and $\beta = 0.1$ would apply \emph{component pruning} twice, resulting in a new netlist with $8$ components. The value $\beta = 0.1$ was determined empirically by analyzing which genetic operations most effectively contributed to elite circuits; this analysis and the resulting choice of $\beta$ are discussed in detail in Sec.~\ref{sec:results:subsec:hyperparameters}.

Note that this operation can sometimes produce invalid netlists if the two component definitions have the same arity (number of terminals and parameters) but completely different semantics (e.g., mixing ``\texttt{M1 D G S B NMOS}'' with ``\texttt{V1 vdd 0 DC 1.8}''). However, such invalid netlists naturally receive low rewards and are not further considered, having only a minor impact on efficiency. Moreover, in all of our experiments, circuits are synthesized only from NMOS/PMOS transistors or from subcircuits, each with a unique arity, so this case does not occur.

%%%%%%%%%%%%%%%%%%%%%%%%%%%%%%%%%%%%%%%%%%%%%%%%%%%%%%%%%%
\squeezeSubSection\subsection{Summary of Full Approach}
\label{sec:spm:subsec:full_approach}
%%%%%%%%%%%%%%%%%%%%%%%%%%%%%%%%%%%%%%%%%%%%%%%%%%%%%%%%%%

First, the population is initialized with the GraCo random sampler, then iteratively refined by one of three operations, each chosen with equal probability:
\begin{compactitem}
    \item Crossover between two elite netlists,
    \item Mutation of an elite netlist with a random netlist,
    \item Pruning to reduce component definitions.
\end{compactitem}
For crossover and mutation, parents are drawn from the elite set using roulette-wheel sampling, where circuits are first ranked by reward and selection probabilities are proportional to this ranking. This method favors high-ranking candidates while preserving diversity by giving lower-ranked ones a chance. Sampling is with replacement, so the same netlist can be chosen twice in crossover.

The elite set size is defined either as a ratio $\eta$ of all evaluated netlists, e.g., $\eta = 0.01$ with $N=1000$ circuits yields $10$ elites that increase over time, or as a fixed size $\zeta$. Sec.~\ref{sec:results:subsec:hyperparameters} shows that $\zeta = 30$ works best for our experiments. Finally, \spm{} outputs the highest-scoring netlist found, which can be visualized using ML-based tools such as the LLM-driven system~\cite{matsuo2024schemato}, helping designers interpret the resulting circuit.

%%%%%%%%%%%%%%%%%%%%%%%%%%%%%%%%%%%%%%%%%%%%%%%%%%%%%%%%%%
\squeezeSection\section{Synthesis Results}
\label{sec:results}
%%%%%%%%%%%%%%%%%%%%%%%%%%%%%%%%%%%%%%%%%%%%%%%%%%%%%%%%%%

We first discuss hyperparameter selection for \spm{}, then present example applications: synthesizing standard digital cells and OpAmps using the open-source Skywater $\qty{130}{\nano\metre}$ PDK~\cite{skywater130pdk}. Please note that these tasks are not limitations of \spm{}, but illustrative case studies. Furthermore, \spm{} is compatible with any PDK, since adapting it only requires replacing the device models in the GraCo library. The genetic operations themselves are PDK-agnostic because they operate directly at the netlist level.

%%%%%%%%%%%%%%%%%%%%%%%%%%%%%%%%%%%%%%%%%%%%%%%%%%%%%%%%%%
\squeezeSubSection\subsection{Optimal \spm{} Hyperparameters}
\label{sec:results:subsec:hyperparameters}
%%%%%%%%%%%%%%%%%%%%%%%%%%%%%%%%%%%%%%%%%%%%%%%%%%%%%%%%%%

\begin{figure*}
    \subfloat[Inverter]{
        \includegraphics[width=0.4\linewidth,trim=15 22 5 20]{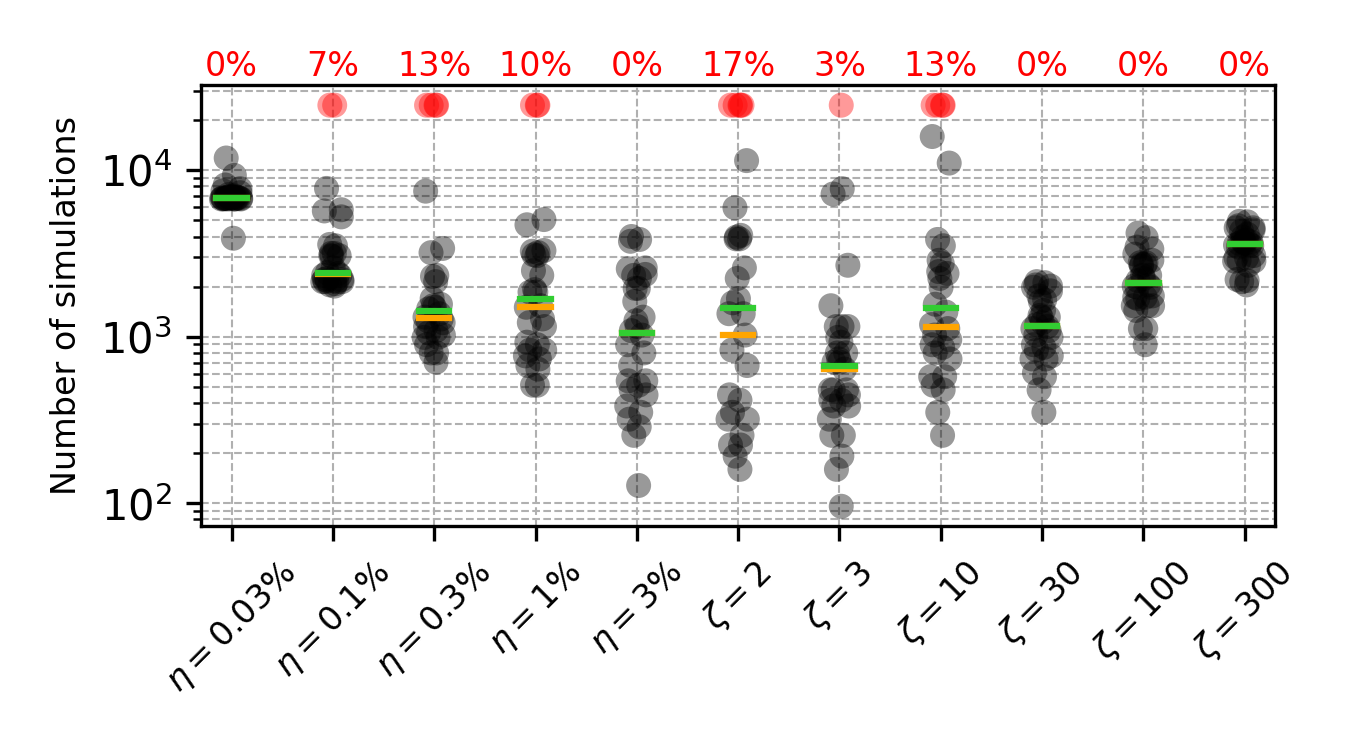}
        \label{fig:inverter_sims}}
    \subfloat[NAND2]{
        \includegraphics[width=0.4\linewidth,trim=15 22 5 20]{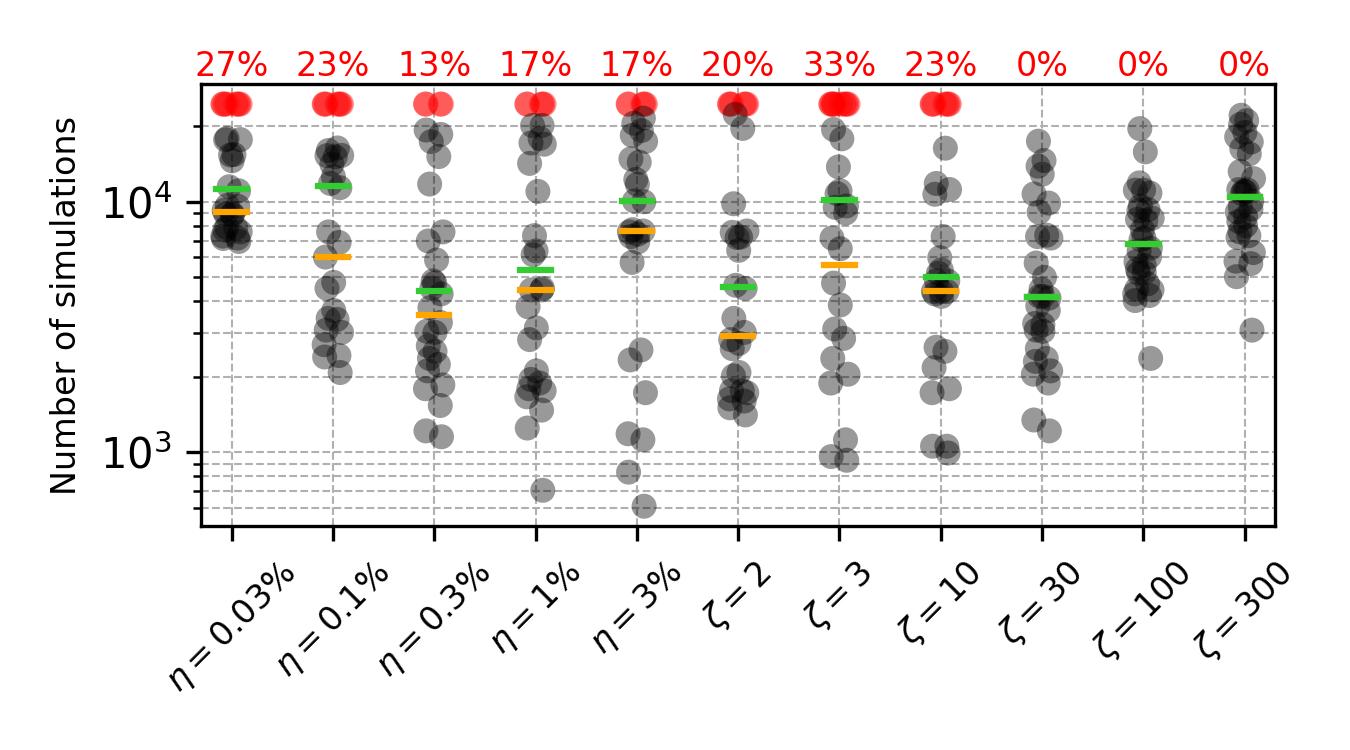}
        \label{fig:nand2_sims}}
    \subfloat[NAND2, w/o normalization]{\raisebox{6pt}{
        \includegraphics[width=0.108\linewidth,trim=0 22 0 20]{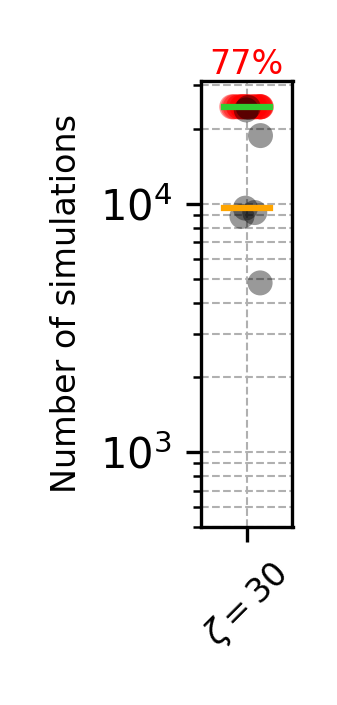}}
        \label{fig:nand2_nonormalization_sims}}
    \caption{SPICE simulations needed to synthesize an inverter and NAND2 for different elite set sizes $\eta$ or $\zeta$. Orange and green lines show median values for successful and all runs (including those hitting the budget of $256 \cdot 32 \cdot 3 \approx 25$k simulations). Red numbers denote failure rates~(\%) where we could not successfully find a circuit. In (c), we present ablation results for the NAND2 gate without applying netlist normalization; see Sec.~\ref{sec:results:subsec:ablation} for the discussion.}
    \Description{SPICE Simulations needed to synthesize an inverter and NAND2 for different elite sizes $\eta$ or $\zeta$. Orange and green lines show median values for successful and all runs (including those hitting the budget of $256 \cdot 32 \cdot 3 \approx 25$k simulations). Red numbers denote failure rates~(\%). In (c), we present ablation results for the NAND2 gate without applying netlist normalization; see Sec.~\ref{sec:results:subsec:ablation} for discussion.}
    \label{fig:inv+nand2_sims}
    \squeezeSpace
\end{figure*}

\begin{figure*}
    \centering
    \includegraphics[width=0.26\linewidth,trim=30 10 30 10]{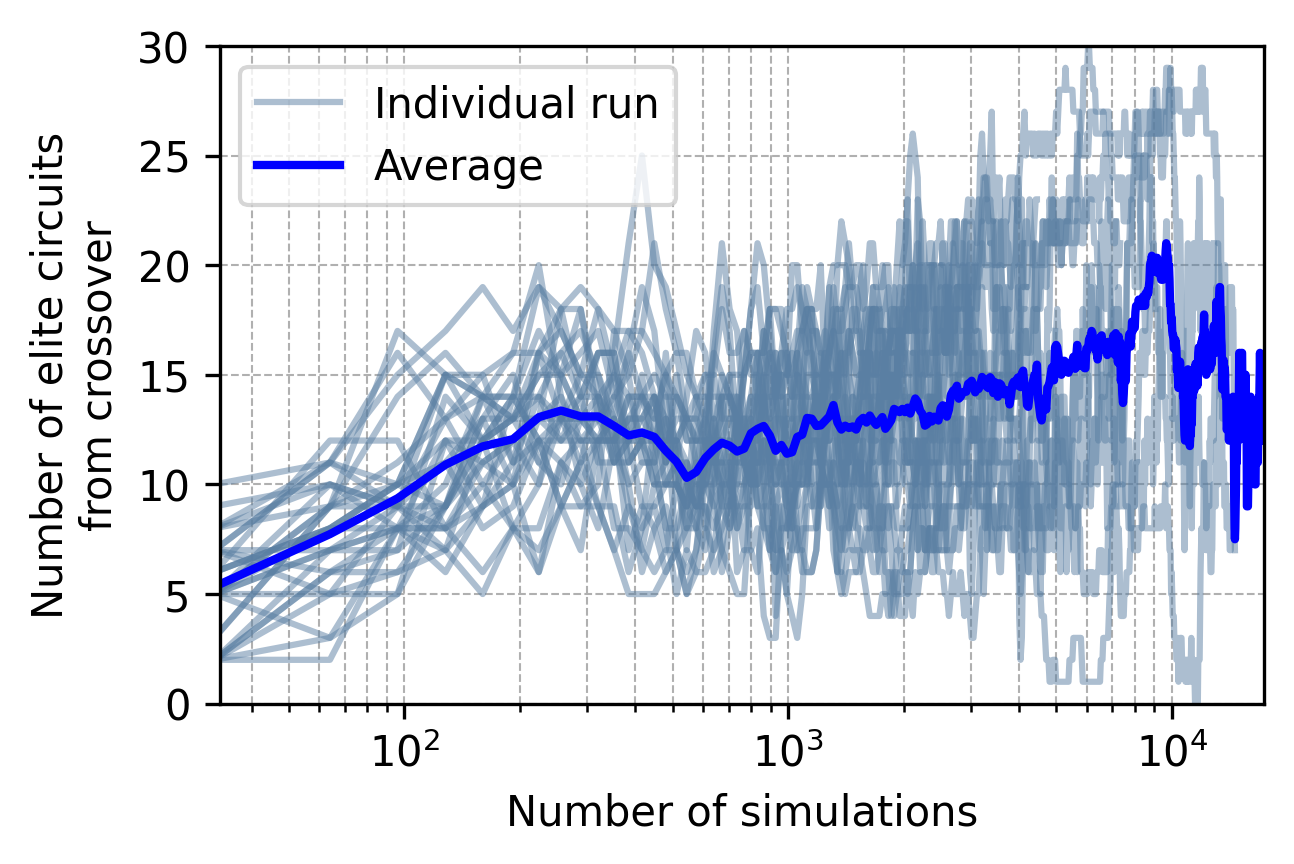}
    \hspace*{1.3cm}
    \includegraphics[width=0.26\linewidth,trim=30 10 30 10]{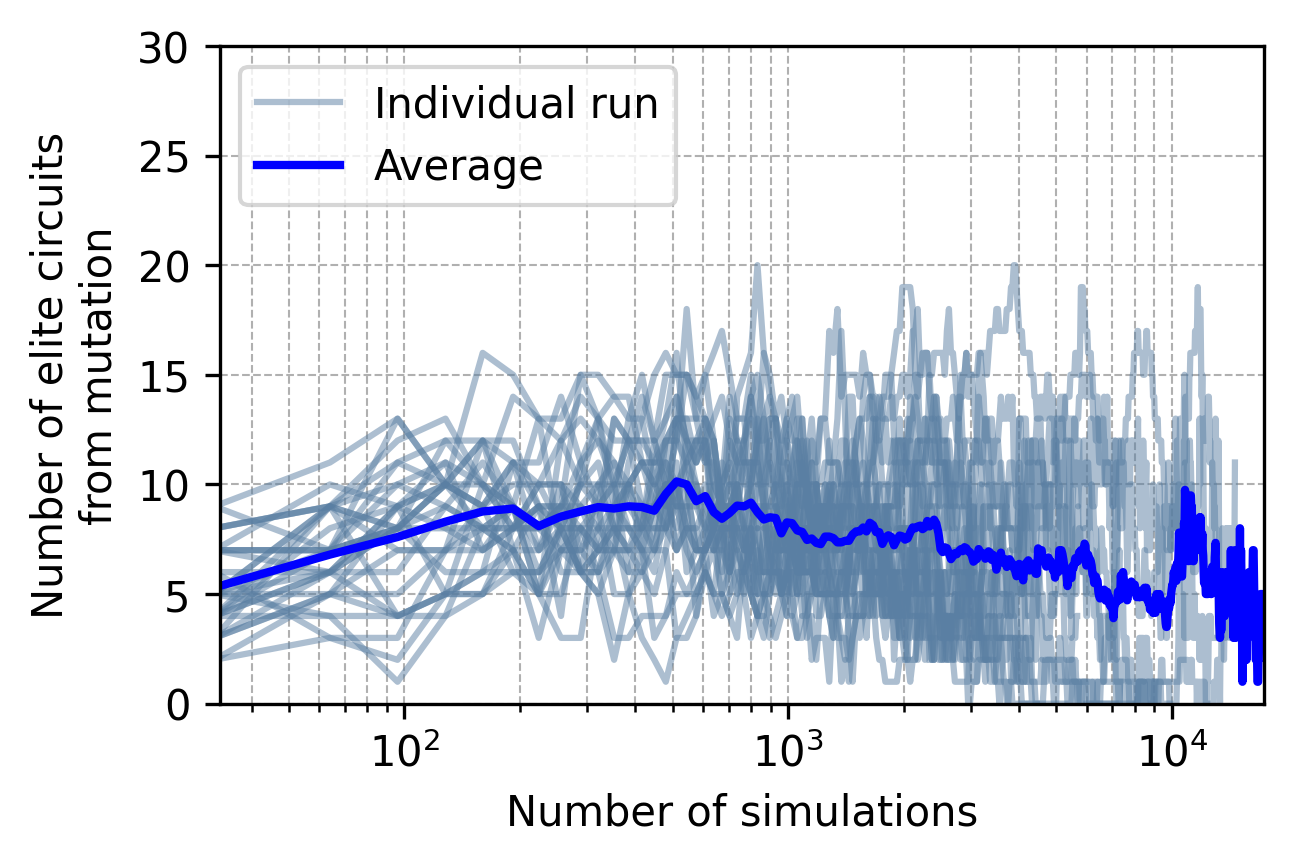}  
    \hspace*{1.3cm}
    \includegraphics[width=0.26\linewidth,trim=30 10 30 10]{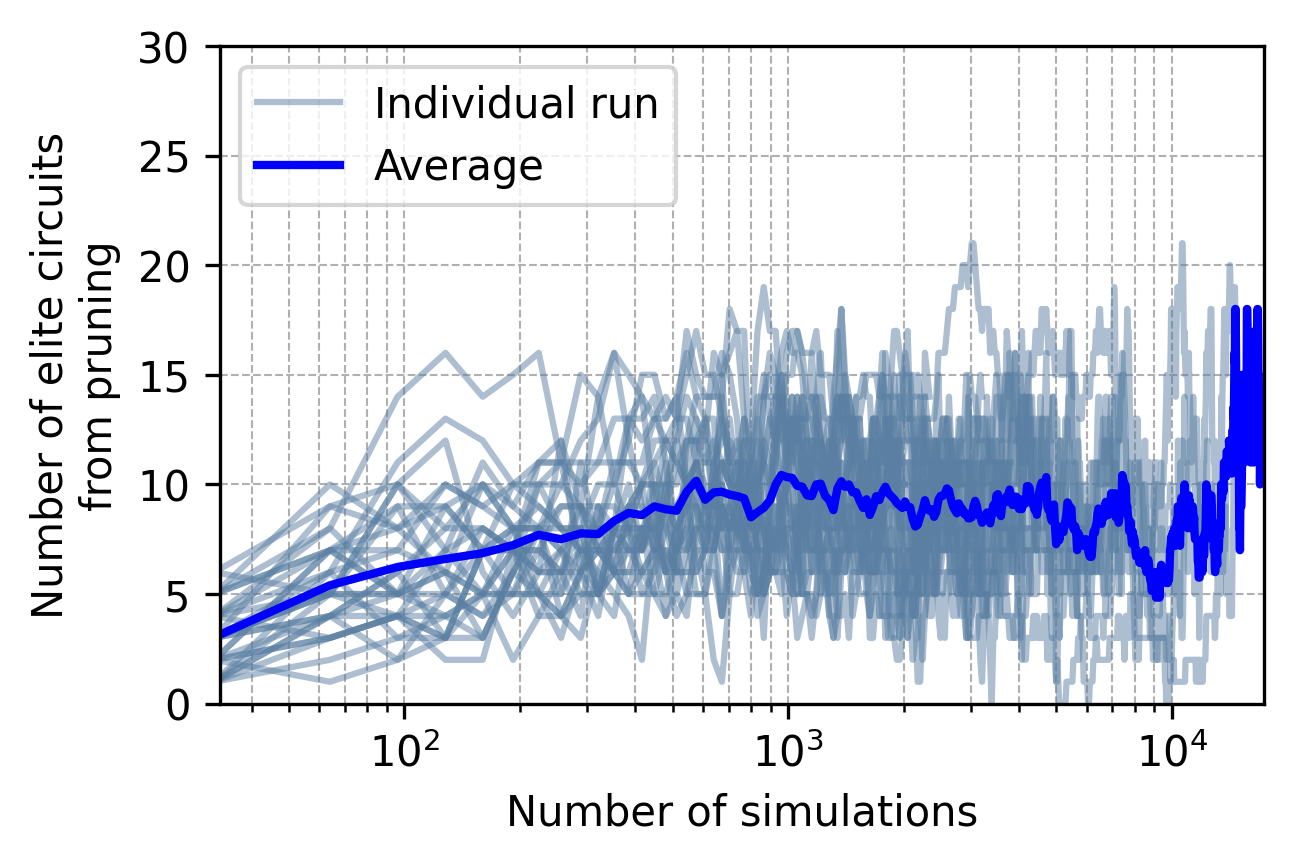}
    \caption{Number of circuits in the elite set ($\zeta = 30$, NAND2 task) generated by a specific genetic operation.}
    \Description{Number of circuits in the elite set ($\zeta = 30$, NAND2 task) generated by a specific genetic operation.}
    \label{fig:nand2_elite_generation}
    \squeezeSpace\squeezeSpace
\end{figure*}

\subsubsection{Elite Set Size ($\eta$, $\zeta$)}
Selecting an appropriate elite set size is crucial, as it strongly affects the performance of a genetic algorithm. For \spm{}, a small elite set leads to insufficient diversity among offspring circuits, causing convergence to suboptimal solutions. Conversely, a large elite set results in poor exploitation of the most successful circuits. To address this, we performed a grid search to identify the optimal values of $\eta$ (for defining a relative elite set size) or $\zeta$ (for defining an absolute elite set size) when synthesizing an inverter and a NAND2 gate. For both tasks, the synthesis process was terminated either when the overall reward, evaluating output voltages and timings through the reward function, reached a value of $1$, indicating successful circuit synthesis, or when the maximum number of simulations was reached. In the latter case the run was counted as a failure.

Fig.~\ref{fig:inv+nand2_sims} shows the number of SPICE simulations required to synthesize an inverter and a NAND2 across $30$ runs. The results show that a fixed elite size of $\zeta = 30$ netlists consistently produces the best outcomes, enabling successful synthesis while minimizing the median number of simulations needed. Using a smaller $\zeta$ reduces diversity within the elite set, which can cause convergence to suboptimal solutions. Conversely, a large $\zeta$ reduces the effectiveness of genetic operations, as more suboptimal netlists are paired together, lowering selection pressure. In addition, using a relative elite size can be detrimental: early in training, when the elite set is small, it may collapse into a degenerate population with limited genetic diversity, from which recovery is difficult.

\squeezeSubSubSection\subsubsection{Mutation Probability $\alpha$ and Pruning Rate $\beta$}
When designing the genetic operations, we introduced two additional hyperparameters: the mutation probability $\alpha$, which controls how many lines from the elite parent are replaced by lines from the random netlist, and the pruning rate $\beta$, which controls how many lines are pruned from the parent netlist. We tuned $\alpha$ and $\beta$ through a preliminary grid search, in which we analyzed how often each genetic operation produced circuits that enter the elite set. Our goal is that all three operations, which are applied with equal probability, have similar ``strengths'', i.e., elite circuits are equally likely to originate from any of them. Using this criterion, we obtain the values $\alpha = 0.3$ and $\beta = 0.1$ given in Sec.~\ref{sec:spm:subsec:genetic_operators}. Fig.~\ref{fig:nand2_elite_generation} shows, for the NAND2 task, the number of elite netlists produced by each genetic operation over 30 runs, together with the corresponding average curves. This analysis confirms that all three operations are similarly effective. Although the GA population is initialized by the GraCo random sampler, these initial circuits are quickly replaced in the elite set by those generated by the genetic operations.

%%%%%%%%%%%%%%%%%%%%%%%%%%%%%%%%%%%%%%%%%%%%%%%%%%%%%%%%%%
\squeezeSubSection\subsection{Standard Cell Synthesis}
\label{sec:results:subsec:standard_cell}
%%%%%%%%%%%%%%%%%%%%%%%%%%%%%%%%%%%%%%%%%%%%%%%%%%%%%%%%%%

\begin{table}
\caption{Effect of consistency checks on NAND2 synthesis with \spm{}. Each method was run three times, with results sorted by value. We report the average train reward (higher is better, averaged over all training steps) and the ``Optimality gap'' $= 1 - \text{Best reward}$ (lower is better). Gap values below $1/15 \approx 0.067$ meet voltage and power constraints but differ in timing, and are shown in \textcolor{blue}{blue}. The \textit{best} and \underline{second-best} circuits are highlighted; on average, the best circuit has a propagation delay that is about $3\,\text{ps}$ smaller than that of the second-best circuit.}
\label{tab:nand2_spicemixer}
\centering
\setlength{\tabcolsep}{3pt}
\resizebox{\linewidth}{!}{
\begin{tabular}{@{}lcccccc@{}}
\toprule
\multirow{2}{*}{\shortstack{\textbf{Consistency Check}\\$\dagger/\ddagger = $ during/after generation}} & \multicolumn{3}{c}{\textbf{Optimality gap}} & \multicolumn{3}{c}{\textbf{Avg. train reward}} \\
\cmidrule(lr){2-4} \cmidrule(lr){5-7}
& Run $1$ & Run $2$ & Run $3$ & Run $1$ & Run $2$ & Run $3$ \\
\midrule
None                                                 & \textcolor{blue}{$6.6\cdot 10^{-7}$} & \textcolor{blue}{$1.2\cdot 10^{-7}$} & \textcolor{blue}{\underline{$6.0\cdot 10^{-8}$}} & 0.79 & 0.87 & 0.89 \\
Connected in/out nets $\dagger$                  & $6.7\cdot 10^{-2}$ & \textcolor{blue}{$3.5\cdot 10^{-6}$} & \textcolor{blue}{$1.8\cdot 10^{-7}$} & 0.79 & 0.82 & 0.87 \\
Paths between in/out nets $\dagger$              & \textcolor{blue}{$5.4\cdot 10^{-7}$} & \textcolor{blue}{$1.2\cdot 10^{-7}$} & \textcolor{blue}{\underline{$6.0\cdot 10^{-8}$}} & 0.74 & 0.78 & 0.88 \\
No floating nets $\dagger$                       & $6.7\cdot 10^{-2}$ & \textcolor{blue}{\underline{$6.0\cdot 10^{-8}$}} & \textcolor{blue}{\underline{$6.0\cdot 10^{-8}$}} & 0.76 & 0.79 & 0.86 \\
No isolated subgraphs $\dagger$                  & $6.7\cdot 10^{-2}$ & \textcolor{blue}{$7.8\cdot 10^{-7}$} & \textcolor{blue}{$1.2\cdot 10^{-7}$} & 0.80 & 0.80 & 0.83 \\
Connected in/out nets $\ddagger$                   & \textcolor{blue}{$1.9\cdot 10^{-4}$} & \textcolor{blue}{$1.2\cdot 10^{-7}$} & \textcolor{blue}{$1.2\cdot 10^{-7}$} & 0.63 & 0.84 & 0.87 \\
Paths between in/out nets $\ddagger$               & \textcolor{blue}{$1.8\cdot 10^{-7}$} & \textcolor{blue}{\underline{$6.0\cdot 10^{-8}$}} & \textcolor{blue}{$\mathit{< 10^{-9}}$} & 0.80 & 0.88 & 0.88 \\
No floating nets $\ddagger$                        & $6.7\cdot 10^{-2}$ & \textcolor{blue}{$1.2\cdot 10^{-7}$} & \textcolor{blue}{$1.2\cdot 10^{-7}$} & 0.79 & 0.84 & 0.88 \\
No isolated subgraphs $\ddagger$                   & $6.7\cdot 10^{-2}$ & \textcolor{blue}{$1.2\cdot 10^{-7}$} & \textcolor{blue}{\underline{$6.0\cdot 10^{-8}$}} & 0.73 & 0.73 & 0.79 \\
All $\dagger$                                    & \textcolor{blue}{$1.2\cdot 10^{-7}$} & \textcolor{blue}{$1.2\cdot 10^{-7}$} & \textcolor{blue}{\underline{$6.0\cdot 10^{-8}$}} & 0.80 & 0.87 & 0.87 \\
All $\ddagger$                                     & \textcolor{blue}{$1.2\cdot 10^{-7}$} & \textcolor{blue}{\underline{$6.0\cdot 10^{-8}$}} & \textcolor{blue}{\underline{$6.0\cdot 10^{-8}$}} & 0.84 & 0.85 & 0.87 \\
\bottomrule
\end{tabular}}
\squeezeSpace
\end{table}
\begin{table}
    \caption{Synthesis of a NAND2 gate with CMA-ES for different random seeds. Circuits with an optimality gap below $1/15 \approx 0.067$ meet voltage and power constraints. Highlighting matches Table~\ref{tab:nand2_spicemixer}.}
    \label{tab:cma_es_nand2}
    \centering
    \setlength{\tabcolsep}{4pt}
    \resizebox{\linewidth}{!}{
    \begin{tabular}{@{}m{2.1cm}m{1.5cm}@{}m{1.5cm}@{}m{1.5cm}@{}m{1.5cm}@{}m{1.27cm}@{}}
    \toprule
    \multirow{2}{*}{\textbf{ Design space}} & \multicolumn{5}{c}{\textbf{Optimality gap ($1-$Best Reward)}} \\
    \cmidrule(lr){2-6}
    & Run 1 & Run 2 & Run 3 & Run 4 & Run 5 \\
    \midrule
    \shortstack{$N_\text{components} = 4$\\[-3pt] $N_\text{internal\_nets} = 1$} & $1.4\cdot 10^{-1}$ & $1.3\cdot 10^{-1}$ & $2.0\cdot 10^{-1}$ & $1.3\cdot 10^{-1}$ & $1.3\cdot 10^{-1}$ \\[8pt]
    \shortstack{$N_\text{components} = 8$\\[-3pt] $N_\text{internal\_nets} = 2$} & $2.2\cdot 10^{-1}$ & \textcolor{blue}{$\mathbf{1.2\cdot 10^{-3}}$} & $2.7\cdot 10^{-1}$ & \underline{$6.7\cdot 10^{-2}$} & $7.4\cdot 10^{-2}$ \\
    \bottomrule
    \end{tabular}}
    \squeezeSpace\squeezeSpace
\end{table}

We now present the results for synthesizing standard cells. For benchmarking, we compare \spm{} with two reinforcement learning (RL) methods from~\cite{uhlich2024graco} (RLOO and ES) and the \emph{Covariance Matrix Adaptation Evolution Strategy} (CMA-ES)~\cite{hansen2016cma}.

%%%%%%%%%%%%%%%%%%%%%%%%%%%%%%%%%%%%%%%%%%%%%%%%%%%%%%%%%%
\squeezeSubSubSection\subsubsection{Inverter}
%%%%%%%%%%%%%%%%%%%%%%%%%%%%%%%%%%%%%%%%%%%%%%%%%%%%%%%%%%

Using the same setup as in~\cite{uhlich2024graco}, we can compare \spm{} with $\zeta = 30$ from Fig.~\ref{fig:inverter_sims} to GraCo~\cite{uhlich2024graco} and find that it consistently produces valid inverter designs, matching GraCo ES in success rate but requiring about $25\times$ fewer simulations, and outperforming GraCo RLOO and the random sampler.

%%%%%%%%%%%%%%%%%%%%%%%%%%%%%%%%%%%%%%%%%%%%%%%%%%%%%%%%%%
\squeezeSubSubSection\subsubsection{NAND2}
%%%%%%%%%%%%%%%%%%%%%%%%%%%%%%%%%%%%%%%%%%%%%%%%%%%%%%%%%%

We evaluate NAND2 synthesis using the same setup as in~\cite{uhlich2024graco}, where saturation applies only to voltage levels and power, not timing, so the synthesizer is asked to find a correct circuit with the fastest timings. Table~\ref{tab:nand2_spicemixer} shows that \spm{} with various consistency checks succeeds in 28 of 33 runs. Here, consistency checks are graph-level structural tests on the circuit that ensure the underlying netlist is a valid SPICE circuit, for example by requiring that input/output nets are connected to devices, that there are paths between inputs and outputs, and that no nets or subcircuits are left floating or isolated. These results far exceed GraCo ES (2/48 successes) and yield circuits that are on average over $200\,\text{ps}$ faster than the best circuit from GraCo ES. The fastest design achieved $T_\text{rise} = \qty{9.45}{\pico\second}$, $T_\text{fall} = \qty{10.35}{\pico\second}$, $T_\text{r2f} = \qty{7.5}{\pico\second}$, and $T_\text{f2r} = \qty{6.93}{\pico\second}$.

We also compare \spm{} with CMA-ES~\cite{hansen2016cma}, a widely used evolutionary strategy that iteratively updates a multivariate Gaussian distribution to balance exploration and exploitation. Each component is encoded by four integer variables (type: \{``unused'', ``Skywater NMOS'', ``Skywater PMOS''\}, drain net, gate net, source net) and two continuous variables (width, length). Manufacturing constraints are the same as in GraCo/\spm{}: the bulk is tied to supply or ground, and supply/ground are not connected to the gate. We use the pycma implementation~\cite{hansen2019pycma} and represent the discrete variables as \texttt{integer\_variables} in pycma to avoid premature collapse of their variance. Table~\ref{tab:cma_es_nand2} summarizes the results. When the number of components and internal nets is fixed to the exact values required for a NAND2 gate, CMA-ES does not synthesize a valid solution. Relaxing these limits by a factor of two improves the outcomes, and in one out of five runs CMA-ES finds a correct circuit. However, these results are still considerably worse than those achieved by \spm{}, underscoring its superiority.

\begin{table}
    \caption{Static latch synthesis comparison. Each method was run four times, with results sorted by train values. A \textcolor{blue}{$\mathbf{0}$} gap indicates passed testbench; N/A indicates failure due to Ngspice convergence issue.}
    \label{tab:latch}
    \centering
    \resizebox{\linewidth}{!}{
    \begin{tabular}{rC{1.6cm}C{1.6cm}C{1.6cm}C{1.6cm}}
    \toprule
    \multirow{2}{*}{\textbf{Synthesis approach}} & \multicolumn{4}{c}{\textbf{Optimality gap ($1 - $Best Reward)}}\\
    \cmidrule(lr){2-5}
    & Run 1 & Run 2 & Run 3 & Run 4 \\
    \midrule
    \multicolumn{5}{l}{\emph{Synthesis testbench (}$t_\text{step} = \qty{1}{\pico\second}$, $t_\text{stop} = \qty{16}{\nano\second}$\emph{)}}\\
    Random & $2.7\cdot 10^{-3}$ & $2.7\cdot 10^{-3}$ & $2.0\cdot 10^{-3}$ & $4.0\cdot 10^{-4}$\\
    GraCo RLOO & $1.1\cdot 10^{-1}$ & $9.2\cdot 10^{-2}$ & $7.3\cdot 10^{-2}$ & $6.8\cdot 10^{-2}$\\
    GraCo ES & $6.7\cdot 10^{-2}$ & $3.7\cdot 10^{-2}$ & $3.2\cdot 10^{-2}$ & $3.0\cdot 10^{-2}$\\
    \spm{} & $1.0\cdot 10^{-4}$ & \textcolor{blue}{{$\mathbf{0}$}} & \textcolor{blue}{{$\mathbf{0}$}} & \textcolor{blue}{{$\mathbf{0}$}}\\
    \\[-5pt]
    \multicolumn{5}{l}{\emph{Verification testbench (}$t_\text{step} = \qty{1}{\nano\second}$, $t_\text{stop} = \qty{1.6}{\milli\second}$\emph{)}}\\
    Random & $5.5\cdot 10^{-2}$ & $5.5\cdot 10^{-2}$ & $4.3\cdot 10^{-2}$ & $2.3\cdot 10^{-2}$\\
    GraCo RLOO & $1.1\cdot 10^{-1}$ & $9.0\cdot 10^{-2}$ & $9.6\cdot 10^{-2}$ & $7.4\cdot 10^{-2}$\\
    GraCo ES & $6.7\cdot 10^{-2}$ & $3.7\cdot 10^{-2}$ & $4.4\cdot 10^{-2}$ & $3.0\cdot 10^{-2}$\\
    \spm{} & N/A & $2.6\cdot 10^{-5}$ & \textcolor{blue}{{$\mathbf{0}$}} & \textcolor{blue}{{$\mathbf{0}$}}\\
    \bottomrule
    \end{tabular}}
    \squeezeSpace
\end{table}

\begin{figure}
    \centering
    \resizebox{0.85\linewidth}{!}{\includegraphics[trim=10 10 10 5]{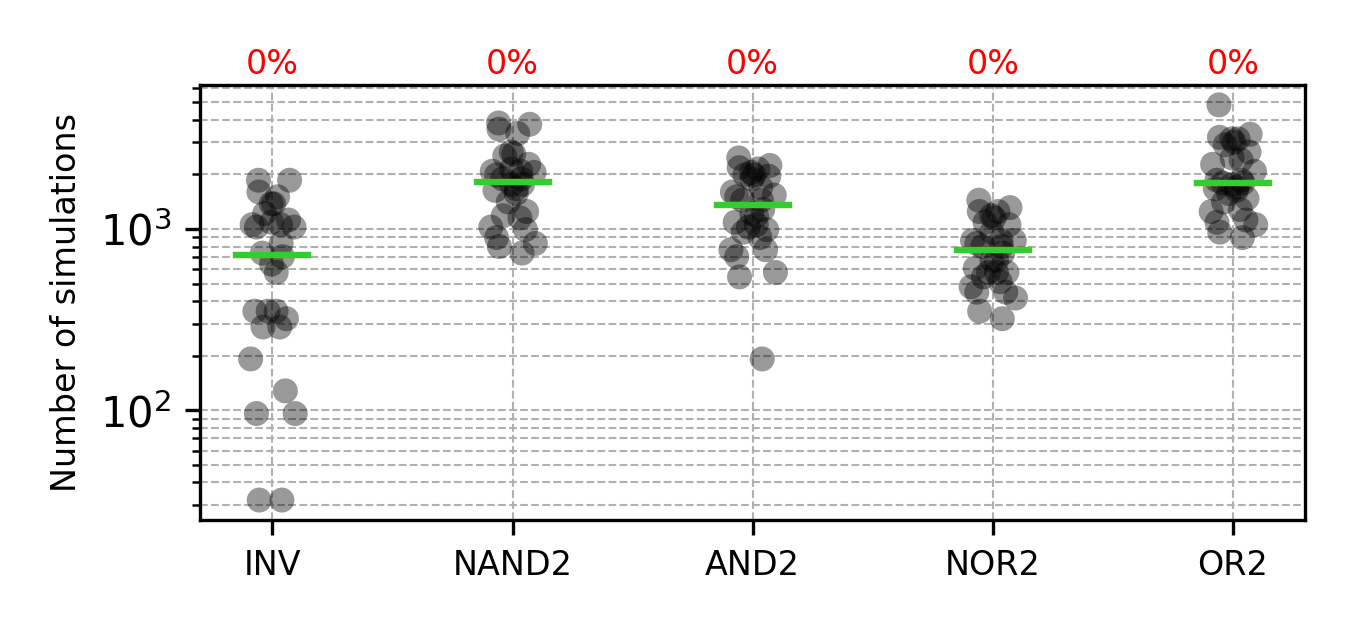}}
    \caption{Simulations for standard cell synthesis. Red numbers denote failure rates (\%) where we could not successfully find a circuit. Inverter and NAND2 require fewer simulations than in Fig.~\ref{fig:inv+nand2_sims} due to fixed transistor sizes.}
    \Description{Simulations for standard cell synthesis. Inverter and NAND2 require fewer simulations than in Fig.~\ref{fig:inv+nand2_sims} due to fixed transistor sizes.}
    \label{fig:scl_synthesis}
    \squeezeSpace\squeezeSpace
\end{figure}

\begin{figure*}[h]
    \vspace{-0.2cm}
    \begin{minipage}[t]{0.64\linewidth}
    \centering
    \subfloat[NMOS-input]{
        \resizebox{0.31\linewidth}{!}{
            \includegraphics[trim=10 10 10 30]{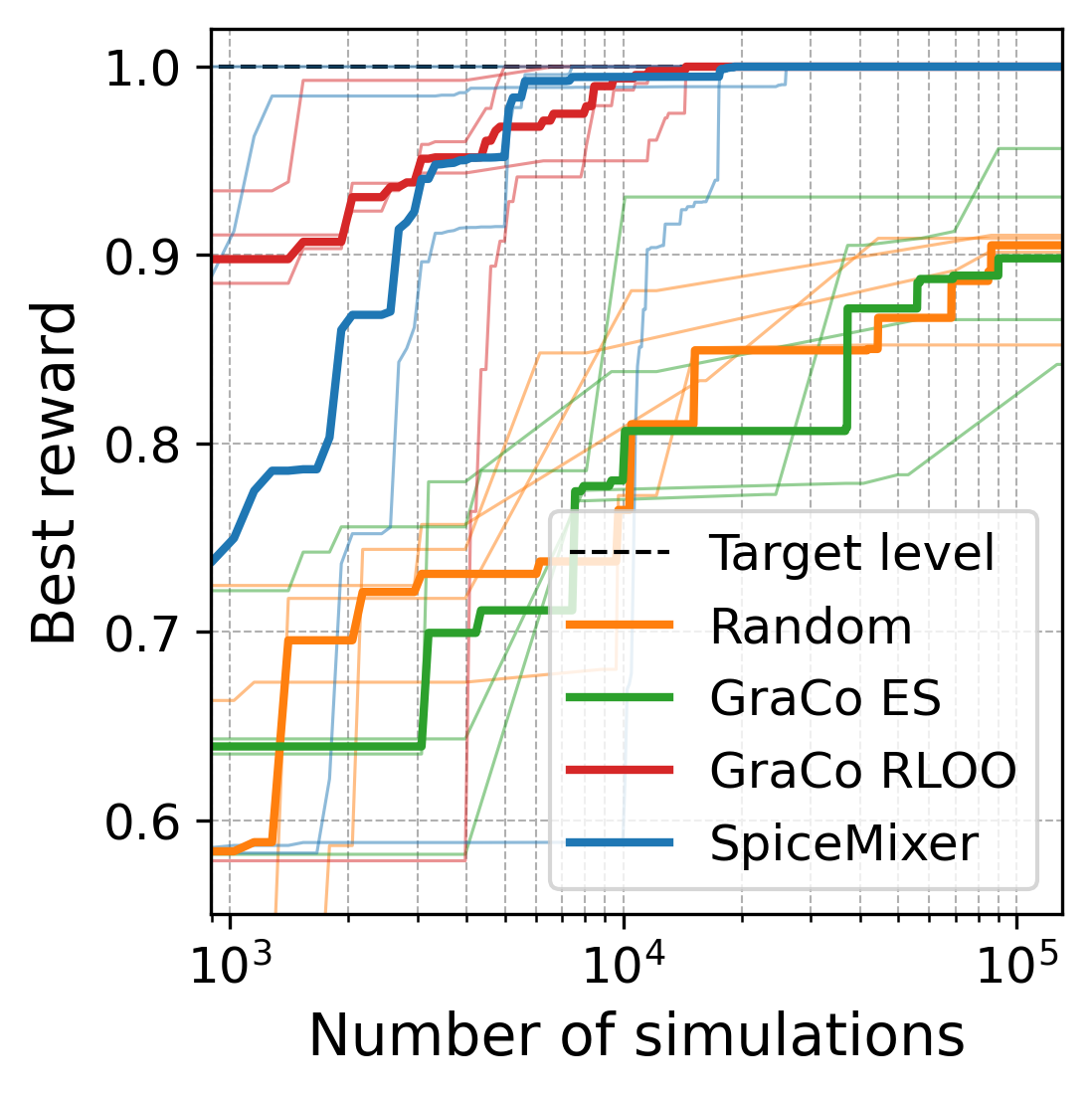}}
    }\hfil
    \subfloat[PMOS-input]{
        \resizebox{0.31\linewidth}{!}{
            \includegraphics[trim=10 10 10 30]{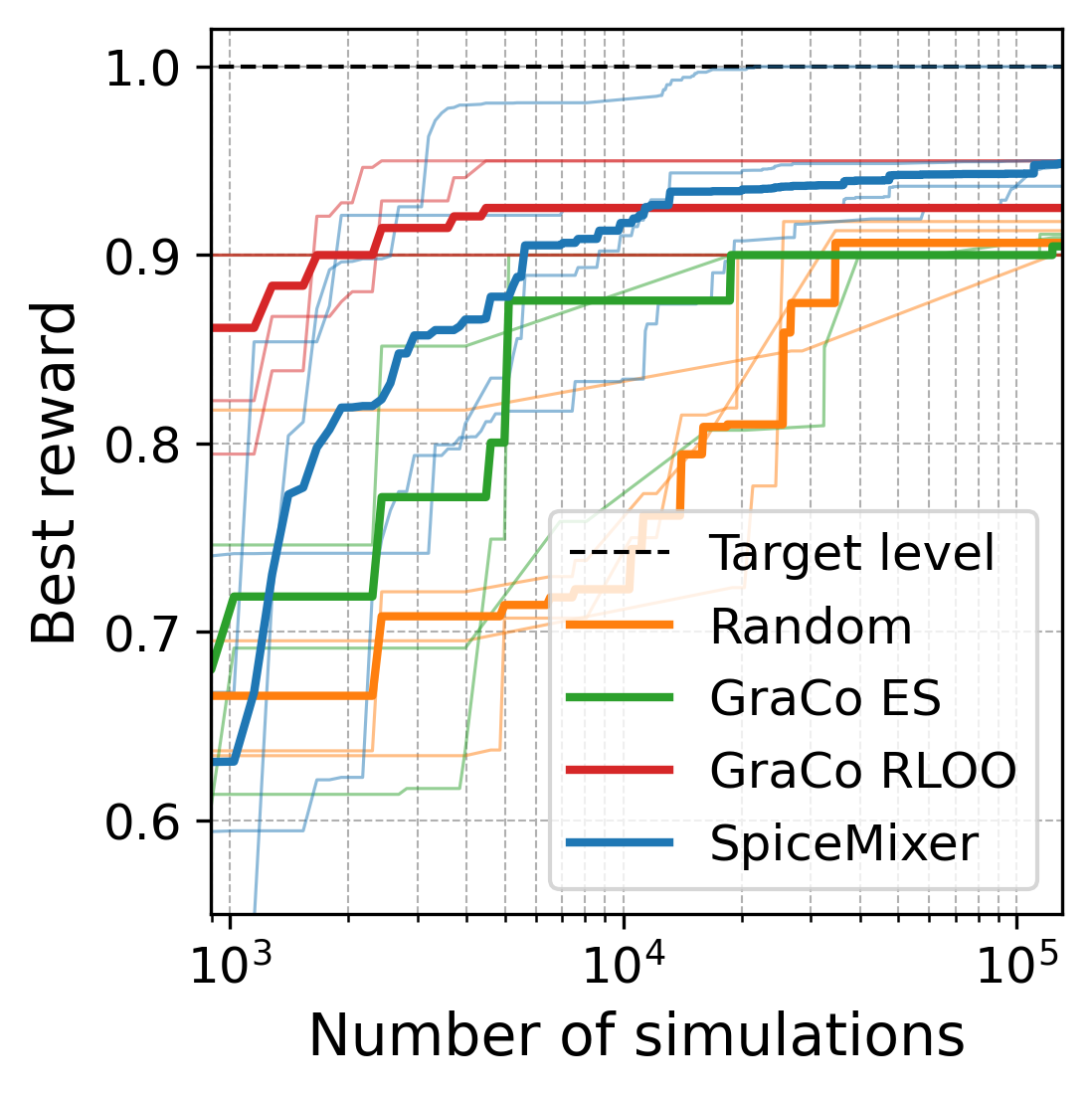}}
    }\hfil
    \subfloat[Complementary-input]{
        \resizebox{0.31\linewidth}{!}{
            \includegraphics[trim=10 10 10 30]{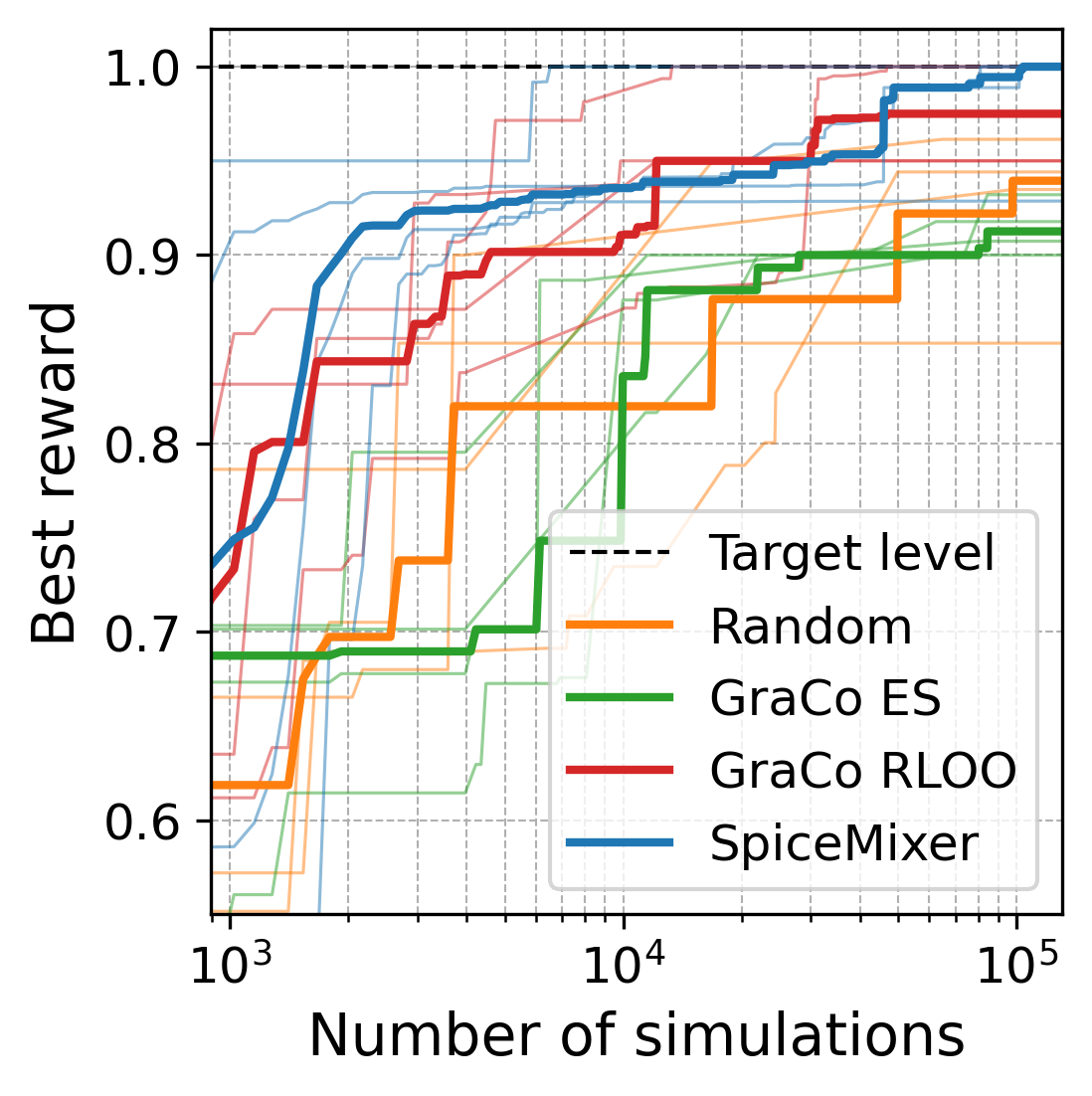}}
    }
    \caption{Evolution of the best reward during OpAmp synthesis. Bold curves represent median over four runs for each method. Reward of $1$ indicates that we found valid circuits passing the testbench and having the specified target metrics.}
    \Description{Evolution of the best reward during OpAmp synthesis. The bold curves represent the median over four runs for each method.}
    \label{fig:opamp_evolution}
    \end{minipage}
    \hfill
    \begin{minipage}[t]{0.33\linewidth}
    \centering
    \resizebox{0.98\linewidth}{!}{
        \includegraphics[trim=50 20 30 50]{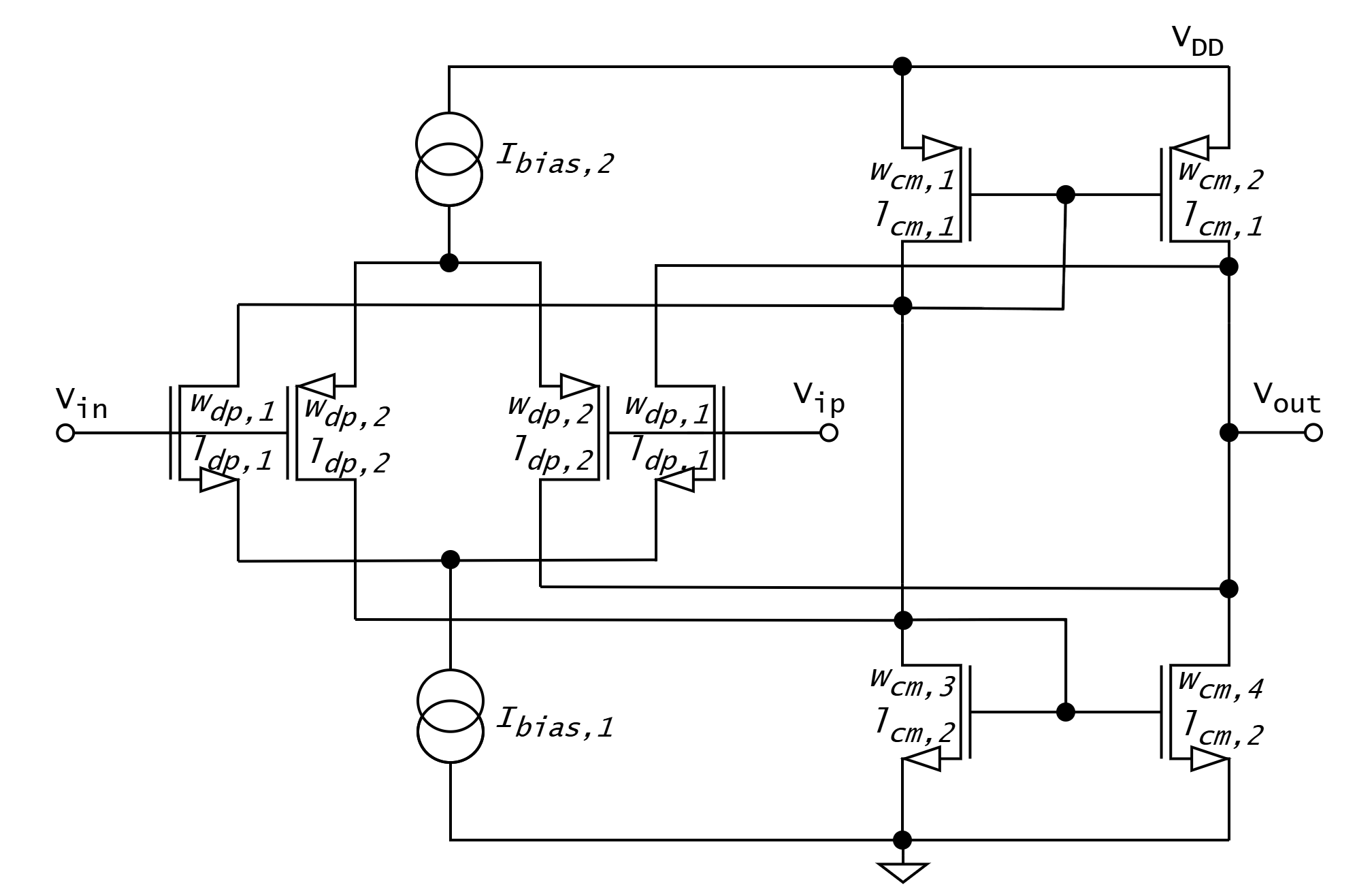}}
    \vspace{0.18cm}
    \caption{Complementary-input OpAmp synthesized by \spm{}. Parameters that needed to be properly sized to achieve target metrics are shown in \textit{italic}.}
    \Description{Complementary-input OpAmp synthesized by \spm{}. Parameters that needed to be properly sized to achieve target metrics are shown in \textit{italic}.}
    \label{fig:opamp_schematics}        
    \end{minipage}
    \squeezeSpace\squeezeSpace
\end{figure*}

%%%%%%%%%%%%%%%%%%%%%%%%%%%%%%%%%%%%%%%%%%%%%%%%%%%%%%%%%%
\squeezeSubSubSection\subsubsection{Latch}
%%%%%%%%%%%%%%%%%%%%%%%%%%%%%%%%%%%%%%%%%%%%%%%%%%%%%%%%%%

As a further standard cell example, we consider a static latch, a level-sensitive storage element transparent when the clock is active, unlike an edge-triggered flip-flop. A static latch retains its state indefinitely while powered, whereas a dynamic latch requires periodic refresh.

The design space consists of two subcircuits: (a) inverters \linebreak\texttt{sky130\_fd\_pr\_\_inv\_01v8} with one input and one output, and (b) tri-state inverters \texttt{sky130\_fd\_pr\_\_invck\_01v8} with two inputs (data, clk) and one output. Transistor dimensions are fixed at $w = \qty{6}{\micro\metre}$, $l = \qty{0.15}{\micro\metre}$, so the task reduces to discovering the correct wiring. Theoretically, one inverter can be used to generate the inverted clock, one tri-state inverter can serve as the input buffer, and a combination of one tri-state inverter and one regular inverter in a cross-coupled loop can provide storage.

Synthesis was tested with a time step $t_\text{step} = \qty{1}{\pico\second}$ and total simulation time $t_\text{stop} = \qty{16}{\nano\second}$, covering $16$ states of $\qty{1}{\nano\second}$ each. As this may only detect dynamic latches, we also ran a longer transient simulation ($t_\text{step} = \qty{1}{\nano\second}$, $t_\text{stop} = \qty{1.6}{\milli\second}$). The reward with saturation enforced correct voltage levels, timings, and power.

From Table~\ref{tab:latch}, \spm{} performed best, producing a static latch in two of four runs, which we also confirmed by manually inspecting the netlists. None of the other methods (Random, GraCo RLOO, GraCo ES) succeeded; notably, the random baseline outperformed both GraCo variants, which collapsed too early to suboptimal solutions.

%%%%%%%%%%%%%%%%%%%%%%%%%%%%%%%%%%%%%%%%%%%%%%%%%%%%%%%%%%
\squeezeSubSubSection\subsubsection{AND2, NOR2, OR2}
%%%%%%%%%%%%%%%%%%%%%%%%%%%%%%%%%%%%%%%%%%%%%%%%%%%%%%%%%%

Finally, we target additional standard cells with fixed transistor sizes, again tasking \spm{} with discovering the correct topology. As shown in Fig.~\ref{fig:scl_synthesis}, \spm{} successfully synthesizes all cells from scratch, requiring on average only roughly $1000$ SPICE evaluations.

\begin{table}[t]
\vspace{0.1cm}
\centering
\caption{Subcircuit library for OpAmp synthesis.}
\label{tab:inventory_opamp}
\resizebox{\linewidth}{!}{%
\begin{tabular}{@{}rcc@{}}
\toprule
\textbf{Type} & \textbf{Subcircuits} & \textbf{Sizing ranges} \\
\midrule
NMOS/PMOS Differential Pair &
\makecell[l]{\shortstack[l]{\Verb|sky130_fd_pr__ndip_01v8|\\\Verb|sky130_fd_pr__pdip_01v8|}}
& $\begin{aligned}
w_\text{dp} &\in [\qty{0.42}{\micro\meter},\, \qty{100}{\micro\meter}]\\
l_\text{dp} &\in [\qty{0.15}{\micro\meter},\, \qty{100}{\micro\meter}]\\
I_\text{bias} &\in [\qty{1}{\micro\ampere},\, \qty{100}{\micro\ampere}]
\end{aligned}$ \\
\\[-5pt]
NMOS/PMOS Current Mirror &
\makecell[l]{\shortstack[l]{\Verb|sky130_fd_pr__ncum_01v8|\\\Verb|sky130_fd_pr__pcum_01v8|}}
& $\begin{aligned}
w_\text{cm,1} &\in [\qty{0.42}{\micro\meter},\, \qty{100}{\micro\meter}]\\
w_\text{cm,2} &\in [\qty{0.42}{\micro\meter},\, \qty{100}{\micro\meter}]\\
l_\text{cm} &\in [\qty{0.15}{\micro\meter},\, \qty{100}{\micro\meter}]\\
\end{aligned}$\\
\bottomrule
\end{tabular}}
\squeezeSpace\squeezeSpace
\end{table}

%%%%%%%%%%%%%%%%%%%%%%%%%%%%%%%%%%%%%%%%%%%%%%%%%%%%%%%%%%
\squeezeSubSection\subsection{Operational Amplifier Synthesis}
\label{sec:results:subsec:opamp}
%%%%%%%%%%%%%%%%%%%%%%%%%%%%%%%%%%%%%%%%%%%%%%%%%%%%%%%%%%

We task \spm{} with synthesizing OpAmps that meet the following design targets: all transistors operate in saturation, the open-loop gain is \qty{\ge 45}{\decibel}, the unity-gain bandwidth is \qty{\ge 1}{\mega\hertz}, the phase margin is \qty{\ge 60}{\degree}, and the gain margin is \qty{\ge 15}{\decibel}. Table~\ref{tab:inventory_opamp} lists the available subcircuits and their corresponding parameter ranges. We consider the case of finding an NMOS-input OpAmp (NMOS differential pair with a PMOS current mirror), a PMOS-input OpAmp (PMOS differential pair with an NMOS current mirror) and a complementary-input OpAmp (all given subcircuits). Note that this task is more difficult than the previous ones because it requires discovering both the wiring (topology) and the device sizes, and OpAmp performance is highly sensitive to sizing~\cite{shi2021sizing}.

Fig.~\ref{fig:opamp_evolution} shows the best reward over four runs per method. For two out of the three cases, \spm{} achieves the highest median reward and thus outperforms the strongest baseline (GraCo RLOO). In particular, it reaches a reward of $1.0$ for the NMOS-input and complementary-input OpAmp tasks, which means that in each of these cases at least three out of four synthesis runs produced a valid OpAmp circuit. The synthesized designs are well-known OpAmp topologies (NMOS/PMOS differential pairs with active loads) and \spm{} sized them to meet all specifications. \spm{} also discovered a complementary-input OpAmp with NMOS and PMOS differential-pair inputs shown in Fig.~\ref{fig:opamp_schematics}. The sized circuit achieves an open-loop gain of \qty{47.85}{\decibel}, a unity-gain bandwidth of \qty{3.72}{\mega\hertz}, a phase margin of \qty{74.41}{\degree}, and a gain margin of \qty{33.09}{\decibel}, with all transistors in saturation matching our target specifications. 

%% NMOS-input
% * Circuit 7
%* sampler: ga, task: opamp_v2, reward: 1.0000 (1.0 (metrics: {'open_loop_gain__db': 47.621368408203125, 'unity_gain_bandwidth__hz': 1548817.0, %'phase_margin__deg': 61.68199920654297, 'gain_margin__db': 24.534799575805664}, transistors in saturation: 10 of 10))
% X2 n0 vout pmos_current_mirror
% X3 vin vip vout n0 sky130_fd_pr__ndip_01v8 nmos_diff_pair
% X4 vip vin n0 vout sky130_fd_pr__ndip_01v8 nmos_diff_pair

%% PMOS-input
% * Circuit 6
% ["* \x1b[1msampler: ga, task: opamp_v2, reward: 1.0000 (1.0 (metrics: {'open_loop_gain__db': 45.25128936767578, 'unity_gain_bandwidth__hz': 16218100.0, 'phase_margin__deg': 72.89900207519531, 'gain_margin__db': 17.638599395751953}, transistors in saturation: 16 of 16))\x1b[22m\n"]X2 n0 vout nmos_current_mirror
% X3 vin vip vout n0 sky130_fd_pr__pdip_01v8 pmos_diff_pair
% X4 vip vin n0 vout sky130_fd_pr__pdip_01v8 pmos_diff_pair

%% Complementary input
% * Circuit 6
% * sampler: ga, task: opamp_v2, reward: 1.0000 (1.0 (metrics: {'open_loop_gain__db': 47.8538703918457, 'unity_gain_bandwidth__hz': 3715352.0, 'phase_margin__deg': 74.40799713134766, 'gain_margin__db': 33.08760070800781}, transistors in saturation: 12 of 12))

% X2 n0 vout nmos_current_mirror
% X3 n0 vout pmos_current_mirror
% X4 vin vip n0 vout sky130_fd_pr__pdip_01v8 pmos_diff_pair
% X5 vip vin n0 vout sky130_fd_pr__ndip_01v8 nmos_diff_pair

%%%%%%%%%%%%%%%%%%%%%%%%%%%%%%%%%%%%%%%%%%%%%%%%%%%%%%%%%%
\squeezeSubSection\subsection{Ablation: Effect of Netlist Normalization}
\label{sec:results:subsec:ablation}
%%%%%%%%%%%%%%%%%%%%%%%%%%%%%%%%%%%%%%%%%%%%%%%%%%%%%%%%%%
To evaluate the effect of netlist normalization, we repeat the NAND2 synthesis from Sec.~\ref{sec:results:subsec:hyperparameters} but disable normalization for all generated netlists. As shown in Fig.~\ref{fig:nand2_nonormalization_sims}, the failure rate increases to 77\% within the given SPICE simulation budget (compared to 0\% with normalization). Without normalization, elite and random netlists differ more in structure, which makes \emph{netlist mixing} and \emph{component mixing} less effective. This observation supports our intuition from Sec.~\ref{sec:spm}.

%%%%%%%%%%%%%%%%%%%%%%%%%%%%%%%%%%%%%%%%%%%%%%%%%%%%%%%%%%
\squeezeSection\section{Conclusions and Outlook}
\label{sec:conclusions}
%%%%%%%%%%%%%%%%%%%%%%%%%%%%%%%%%%%%%%%%%%%%%%%%%%%%%%%%%%

In this work, we introduced \spm{}, a novel genetic algorithm framework for analog circuit synthesis that operates directly on normalized SPICE netlists. By applying netlist-level genetic operations (crossover, mutation, and pruning), our method effectively explores the design space and consistently outperforms prior approaches such as GraCo and CMA-ES in synthesizing standard cells or an OpAmp.

Looking ahead, we see several directions to improve \spm{}. First, we want to better control the growth of the netlist during synthesis, where circuits can accumulate redundant lines. Often, these extra lines only adjust effective sizing, e.g., parallel MOSFETs can act as a single multi-finger device. Beyond simple constraints on component counts or reward penalties, as explored in~\cite{campilho2024analog}, we could introduce specialized pruning operations that merge such redundant lines while adapting the sizing. Second, enhancing the genetic operations themselves by incorporating more sophisticated, domain-aware crossover and mutation mechanisms could further improve both search efficiency and solution quality. Overall, \spm{} provides a scalable, automated path for analog design, and we believe these improvements will push its capabilities even further.

\pagebreak

%%%%%%%%%%%%%%%%%%%%%%%%%%%%%%%%%%%%%%%%%%%%%%%%%%%%%%%%%%
\bibliographystyle{ACM-Reference-Format}
\bibliography{references}
%%%%%%%%%%%%%%%%%%%%%%%%%%%%%%%%%%%%%%%%%%%%%%%%%%%%%%%%%%

\ifthenelse{\boolean{addappendix}}{

\clearpage

%%%%%%%%%%%%%%%%%%%%%%%%%%%%%%%%%%%%%%%%%%%%%%%%%%%%%%%%%%
\appendix
%%%%%%%%%%%%%%%%%%%%%%%%%%%%%%%%%%%%%%%%%%%%%%%%%%%%%%%%%%

The following algorithms provide further details about \spm{}. Alg.~\ref{alg:netlist_mixing} illustrates the process of mixing two netlists for crossover. Algs.~\ref{alg:component_mixing} and \ref{alg:pruning} describe the implementation of the pruning genetic operation. Lastly, Alg.~\ref{alg:netlist_generation} summarizes the overall workflow of \spm{}.

\input{alg_netlist_merging}
\newpage

\input{alg_component_merging}

\clearpage

\input{alg_pruning}
\vspace*{3cm}

\input{alg_genetic_algorithm}

\clearpage
}

\end{document}

%% file: example_netlist_mixing.tex
% crossover
% parents [tensor(0.3610), tensor(0.2865)] offspring tensor(0.7450) ['X0 \textcolor{supplycolor}{0} net_input_0 net_internal_0 net_supply_0 sky130_fd_pr__pfet_01v8 w=1.3300000429153442 l=1.1699999570846558\nX1 net_output_0 net_input_0 net_supply_0 net_supply_0 sky130_fd_pr__pfet_01v8 w=4.170000076293945 l=4.260000228881836\nX2 net_internal_1 net_internal_1 net_output_0 net_supply_0 sky130_fd_pr__pfet_01v8 w=2.2200000286102295 l=0.5519999861717224\n', 'X0 net_internal_0 net_input_0 net_output_0 net_supply_0 sky130_fd_pr__pfet_01v8 w=3.190000057220459 l=0.1599999964237213\nX1 net_output_0 net_input_1 net_supply_0 net_supply_0 sky130_fd_pr__pfet_01v8 w=4.539999961853027 l=2.309999942779541\nX2 \textcolor{supplycolor}{0} net_internal_1 net_output_0 net_supply_0 sky130_fd_pr__pfet_01v8 w=1.0499999523162842 l=1.6399999856948853\nX3 net_internal_0 net_internal_1 net_output_0 \textcolor{supplycolor}{0} sky130_fd_pr__nfet_01v8 w=1.399999976158142 l=0.6970000267028809\nX4 net_internal_2 net_internal_1 net_output_0 \textcolor{supplycolor}{0} sky130_fd_pr__nfet_01v8 w=1.149999976158142 l=0.5709999799728394\n'] X0 net_output_0 net_input_0 net_supply_0 net_supply_0 sky130_fd_pr__pfet_01v8 w=4.170000076293945 l=4.260000228881836
% X1 net_output_0 net_input_1 net_supply_0 net_supply_0 sky130_fd_pr__pfet_01v8 w=4.539999961853027 l=2.309999942779541
% X2 net_internal_0 net_internal_1 net_output_0 \textcolor{supplycolor}{0} sky130_fd_pr__nfet_01v8 w=1.149999976158142 l=0.5709999799728394

\begin{figure*}
    \squeezeSpace\squeezeSpace
    \centering
    \begin{prompt}[Elite parent 1, Reward = 0.3610]
        \fontsize{6pt}{6pt}\selectfont
        \ttfamily\hspace*{1cm}
        \begin{tabular}{lm{1.4cm}m{1.4cm}m{1.4cm}m{1.4cm}ll}
            X0 & \textcolor{supplycolor}{0} & \textcolor{inputcolor}{net\_input\_0} & \textcolor{internalcolor}{net\_internal\_0} & \textcolor{supplycolor}{net\_supply\_0} & sky130\_fd\_pr\_\_pfet\_01v8 w=1.330 l=1.170 \\
            X1 & \textcolor{outputcolor}{net\_output\_0} & \textcolor{inputcolor}{net\_input\_0} & \textcolor{supplycolor}{net\_supply\_0} & \textcolor{supplycolor}{net\_supply\_0} & sky130\_fd\_pr\_\_pfet\_01v8 w=4.170 l=4.260 \\
            X2 & \textcolor{internalcolor}{net\_internal\_1} & \textcolor{internalcolor}{net\_internal\_1} & \textcolor{outputcolor}{net\_output\_0} & \textcolor{supplycolor}{net\_supply\_0} & sky130\_fd\_pr\_\_pfet\_01v8 w=2.220 l=0.552
        \end{tabular}
    \end{prompt}
    \vspace{-0.225cm}
    \begin{prompt}[Elite parent 2, Reward = 0.2865]
        \fontsize{6pt}{6pt}\selectfont
        \ttfamily\hspace*{1cm}
        \begin{tabular}{lm{1.4cm}m{1.4cm}m{1.4cm}m{1.4cm}ll}
            X0 & \textcolor{internalcolor}{net\_internal\_0} & \textcolor{inputcolor}{net\_input\_0} & \textcolor{outputcolor}{net\_output\_0} & \textcolor{supplycolor}{net\_supply\_0} & sky130\_fd\_pr\_\_pfet\_01v8 w=3.190 l=0.160 \\
            X1 & \textcolor{outputcolor}{net\_output\_0} & \textcolor{inputcolor}{net\_input\_1} & \textcolor{supplycolor}{net\_supply\_0} & \textcolor{supplycolor}{net\_supply\_0} & sky130\_fd\_pr\_\_pfet\_01v8 w=4.540 l=2.310 \\
            X2 & \textcolor{supplycolor}{0} & \textcolor{internalcolor}{net\_internal\_1} & \textcolor{outputcolor}{net\_output\_0} & \textcolor{supplycolor}{net\_supply\_0} & sky130\_fd\_pr\_\_pfet\_01v8 w=1.050 l=1.640 \\
            X3 & \textcolor{internalcolor}{net\_internal\_0} & \textcolor{internalcolor}{net\_internal\_1} & \textcolor{outputcolor}{net\_output\_0} & \textcolor{supplycolor}{0} & sky130\_fd\_pr\_\_nfet\_01v8 w=1.400 l=0.697 \\
            X4 & \textcolor{internalcolor}{net\_internal\_2} & \textcolor{internalcolor}{net\_internal\_1} & \textcolor{outputcolor}{net\_output\_0} & \textcolor{supplycolor}{0} & sky130\_fd\_pr\_\_nfet\_01v8 w=1.150 l=0.571
        \end{tabular}
    \end{prompt}
    \vspace{-0.225cm}
    \begin{prompt}[Offspring (after normalization), Reward = 0.7450]
        \fontsize{6pt}{6pt}\selectfont
        \ttfamily\hspace*{1cm}
        \begin{tabular}{lm{1.4cm}m{1.4cm}m{1.4cm}m{1.4cm}ll}
            X0 & \textcolor{outputcolor}{net\_output\_0} & \textcolor{inputcolor}{net\_input\_0} & \textcolor{supplycolor}{net\_supply\_0} & \textcolor{supplycolor}{net\_supply\_0} & sky130\_fd\_pr\_\_pfet\_01v8 w=4.170 l=4.260 & \textcolor{gray}{\textsf{\emph{`X1` from parent 1}}}\\
            X1 & \textcolor{outputcolor}{net\_output\_0} & \textcolor{inputcolor}{net\_input\_1} & \textcolor{supplycolor}{net\_supply\_0} & \textcolor{supplycolor}{net\_supply\_0} & sky130\_fd\_pr\_\_pfet\_01v8 w=4.540 l=2.310  & \textcolor{gray}{\textsf{\emph{`X1` from parent 2}}}\\
            X2 & \textcolor{internalcolor}{net\_internal\_0} & \textcolor{internalcolor}{net\_internal\_1} & \textcolor{outputcolor}{net\_output\_0} & \textcolor{supplycolor}{0} & sky130\_fd\_pr\_\_nfet\_01v8 w=1.150 l=0.571  & \textcolor{gray}{\textsf{\emph{`X4` from parent 2}}}
        \end{tabular}
    \end{prompt}
    \vspace*{-0.15cm}
    \caption{Example of crossover using \emph{netlist mixing} for the NAND2 task. Consistent net naming (\textcolor{inputcolor}{input}, \textcolor{outputcolor}{output}, \textcolor{internalcolor}{internal}, \textcolor{supplycolor}{supply}) enables meaningful merging of parent netlists, yielding an offspring with higher reward than both parents.}
    \Description{Example of crossover using \emph{netlist mixing}. Consistent net naming (\textcolor{inputcolor}{input}, \textcolor{outputcolor}{output}, \textcolor{internalcolor}{internal}, \textcolor{supplycolor}{supply}) enables meaningful merging of parent netlists, yielding an offspring with higher reward than both parents.}
    \label{fig:example_netlist_mixing}
    \squeezeSpace\squeezeSpace
\end{figure*}

%% file: example_netlist_pruning.tex
\begin{figure*}
    \centering
    \begin{prompt}[Elite parent, Reward = 0.5257]
        \fontsize{6pt}{6pt}\selectfont
        \ttfamily\hspace*{1cm}
        \begin{tabular}{lm{1.4cm}m{1.4cm}m{1.4cm}m{1.4cm}ll}
            X0 & \textcolor{supplycolor}{0} & \textcolor{inputcolor}{net\_input\_0} & \textcolor{internalcolor}{net\_internal\_0} & \textcolor{supplycolor}{0} & sky130\_fd\_pr\_\_nfet\_01v8 w=0.741 l=0.205 \\
            X1 & \textcolor{supplycolor}{0} & \textcolor{inputcolor}{net\_input\_0} & \textcolor{internalcolor}{net\_internal\_0} & \textcolor{supplycolor}{0} & sky130\_fd\_pr\_\_nfet\_01v8 w=0.741 l=0.205 \\
            X2 & \textcolor{internalcolor}{net\_internal\_0} & \textcolor{inputcolor}{net\_input\_1} & \textcolor{outputcolor}{net\_output\_0} & \textcolor{supplycolor}{0} & sky130\_fd\_pr\_\_nfet\_01v8 w=0.440 l=2.450 \\
            X3 & \textcolor{outputcolor}{net\_output\_0} & \textcolor{inputcolor}{net\_input\_0} & \textcolor{supplycolor}{net\_supply\_0} & \textcolor{supplycolor}{net\_supply\_0} & sky130\_fd\_pr\_\_pfet\_01v8 w=1.340 l=1.680 \\
            X4 & \textcolor{outputcolor}{net\_output\_0} & \textcolor{inputcolor}{net\_input\_0} & \textcolor{supplycolor}{net\_supply\_0} & \textcolor{supplycolor}{net\_supply\_0} & sky130\_fd\_pr\_\_pfet\_01v8 w=1.340 l=1.680 \\
            X5 & \textcolor{supplycolor}{0} & \textcolor{internalcolor}{net\_internal\_0} & \textcolor{internalcolor}{net\_internal\_1} & \textcolor{supplycolor}{0} & sky130\_fd\_pr\_\_nfet\_01v8 w=0.922 l=0.529 \\
            X6 & \textcolor{supplycolor}{0} & \textcolor{internalcolor}{net\_internal\_0} & \textcolor{internalcolor}{net\_internal\_1} & \textcolor{supplycolor}{0} & sky130\_fd\_pr\_\_nfet\_01v8 w=0.976 l=0.726 \\
            X7 & \textcolor{outputcolor}{net\_output\_0} & \textcolor{internalcolor}{net\_internal\_2} & \textcolor{supplycolor}{net\_supply\_0} & \textcolor{supplycolor}{net\_supply\_0} & sky130\_fd\_pr\_\_pfet\_01v8 w=3.250 l=0.163
        \end{tabular}
    \end{prompt}
    \vspace{-0.225cm}
    \begin{prompt}[Offspring (after normalization), Reward = 0.9280]
        \fontsize{6pt}{6pt}\selectfont
        \ttfamily\hspace*{1cm}
        \begin{tabular}{lm{1.4cm}m{1.4cm}m{1.4cm}m{1.4cm}ll}
            X0 & \textcolor{supplycolor}{0} & \textcolor{inputcolor}{net\_input\_0} & \textcolor{internalcolor}{net\_internal\_0} & \textcolor{supplycolor}{0} & sky130\_fd\_pr\_\_nfet\_01v8 w=0.741 l=0.205 & \textcolor{gray}{\textsf{\emph{`X0` from parent}}}\\
            X1 & \textcolor{supplycolor}{0} & \textcolor{inputcolor}{net\_input\_0} & \textcolor{internalcolor}{net\_internal\_0} & \textcolor{supplycolor}{0} & sky130\_fd\_pr\_\_nfet\_01v8 w=0.741 l=0.205 & \textcolor{gray}{\textsf{\emph{`X1` from parent}}}\\
            X2 & \textcolor{internalcolor}{net\_internal\_0} & \textcolor{inputcolor}{net\_input\_1} & \textcolor{outputcolor}{net\_output\_0} & \textcolor{supplycolor}{0} & sky130\_fd\_pr\_\_nfet\_01v8 w=0.440 l=2.450 & \textcolor{gray}{\textsf{\emph{`X2` from parent}}}\\
            X3 & \textcolor{outputcolor}{net\_output\_0} & \textcolor{inputcolor}{net\_input\_0} & \textcolor{supplycolor}{net\_supply\_0} & \textcolor{supplycolor}{net\_supply\_0} & sky130\_fd\_pr\_\_pfet\_01v8 w=1.340 l=1.680 & \textcolor{gray}{\textsf{\emph{`X3` from parent}}}\\
            X4 & \textcolor{supplycolor}{0} & \textcolor{internalcolor}{net\_internal\_0} & \textcolor{internalcolor}{net\_internal\_1} & \textcolor{supplycolor}{0} & sky130\_fd\_pr\_\_nfet\_01v8 w=0.976 l=0.726 & \textcolor{gray}{\textsf{\emph{`X6` from parent}}}\\
            X5 & \textcolor{outputcolor}{net\_output\_0} & \textcolor{internalcolor}{net\_internal\_0} & \textcolor{supplycolor}{net\_supply\_0} & \textcolor{supplycolor}{0} & sky130\_fd\_pr\_\_nfet\_01v8 w=1.340 l=1.680 & \textcolor{gray}{\textsf{\emph{`X4` and `X5` mixed from parent}}}\\
            X6 & \textcolor{outputcolor}{net\_output\_0} & \textcolor{internalcolor}{net\_internal\_2} & \textcolor{supplycolor}{net\_supply\_0} & \textcolor{supplycolor}{net\_supply\_0} & sky130\_fd\_pr\_\_pfet\_01v8 w=3.250 l=0.163 & \textcolor{gray}{\textsf{\emph{`X7` from parent}}}
        \end{tabular}
    \end{prompt}
    \vspace*{-0.15cm}
    \caption{Example of pruning via \emph{component mixing} for the NAND2 task. PMOS \texttt{X4} and NMOS \texttt{X5} from the parent are merged into a new NMOS, inserted as \texttt{X5} after netlist normalization. The resulting offspring achieves a higher reward.}
    \Description{Example of pruning via \emph{component mixing}. PMOS \texttt{X4} and NMOS \texttt{X5} from the parent are merged into a new NMOS, inserted as \texttt{X5} after netlist normalization. The resulting offspring achieves a higher reward.}
    \label{fig:example_component_mixing}
    \squeezeSpace\squeezeSpace
\end{figure*}

%% file: alg_netlist_merging.tex
\begin{algorithm}
\footnotesize
\DontPrintSemicolon
\SetAlgoNlRelativeSize{-1}       
\SetNlSty{}{}{:}    
\SetKwComment{Comment}{\textcolor{gray}{\#}$\,$}{}
\SetKwComment{tcc}{\textcolor{gray}{\#}$\,$}{}
\SetKwComment{tcp}{\textcolor{gray}{\#}$\,$}{}
\SetKwComment{docsign}{\textcolor{orange}{\texttt{"""}}}{}
\SetKwComment{doctext}{}{}
\SetKwFunction{FMergeNetlists}{MixNetlists}
\SetKwFunction{FSplit}{split}
\SetKwFunction{FZipLongest}{zip\_longest}
\SetKwFunction{FRandomChoice}{RandomChoice}
\SetKwFunction{FAppend}{append}
\SetKwFunction{FEnumerate}{enumerate}
\SetKwFunction{FReSub}{re.sub}
\SetKwFunction{FJoin}{join}
\SetKwProg{Fn}{Function}{}{}

\Fn{\FMergeNetlists{parent1: str, parent2: str}}{
    \docsign{}
    \doctext{\rmfamily\itshape\textcolor{orange}{Mix two netlists into a new offspring.}}
    \doctext{}
    \doctext{\rmfamily\itshape\textcolor{orange}{Note: The following code is for mixing two elite netlists, where lines are uniformly selected from each. With mutation, we prioritize keeping lines from the elite netlist, modifying only 30\% of its lines during mixing.}}
    \docsign{}

    \textbf{\textit{\textcolor{darkgreen}{Initialization}}} \;
    offspring $\leftarrow$ [ ] \Comment*[r]{\rmfamily\itshape\textcolor{gray}{initialize empty offspring list}}

    parent1 $\leftarrow$ \FSplit{parent1, \str{\textbackslash n}} \Comment*[r]{\rmfamily\itshape\textcolor{gray}{split into lines}}
    
    parent2 $\leftarrow$ \FSplit{parent2, \str{\textbackslash n}} \Comment*[r]{\rmfamily\itshape\textcolor{gray}{split into lines}}

    \BlankLine
    \textbf{\textit{\textcolor{darkgreen}{Merge lines from both netlists}}} \;

    \ForEach{(line1, line2) in \FZipLongest{parent1, parent2}}{

        \tcc{\rmfamily\itshape\textcolor{gray}{randomly choose action for line (equal prob.)}}
        action $\leftarrow$ \FRandomChoice{\newline \str{first}, \str{second}, \str{both}, \str{none}} 

        \BlankLine
        
        \If{action $==$ \str{first}}{
            \tcc{\rmfamily\itshape\textcolor{gray}{append line1 from parent1 (if not yet exhausted)}}
            \If{line1 \textbf{\emph{is not}} \emph{None}}{
                offspring.\FAppend{line1} 
            }   
        }
        \ElseIf{action $==$ \str{second}}{
            \tcc{\rmfamily\itshape\textcolor{gray}{append line2 from parent2 (if not yet exhausted)}}
            \If{line2 \textbf{\emph{is not}} \emph{None}}{
                offspring.\FAppend{line2} 
            }
                
        }
        \ElseIf{action $==$ \str{both}}{
            \tcc{\rmfamily\itshape\textcolor{gray}{append line1 from parent1 (if not yet exhausted)}}
            \If{line1 \textbf{\emph{is not}} \emph{None}}{
                offspring.\FAppend{line1} 
            }
            \tcc{\rmfamily\itshape\textcolor{gray}{append line2 from parent2 (if not yet exhausted)}}
            \If{line2 \textbf{\emph{is not}} \emph{None}}{
                offspring.\FAppend{line2}
            }
        }
        \Else{
            \textbf{continue}
            \tcp{\rmfamily\itshape\textcolor{gray}{skip both lines}}

        }
    }

    \BlankLine
    \textbf{\textit{\textcolor{darkgreen}{Renumber components to avoid name conflicts}}} \;
    \ForEach{(index, line) in \FEnumerate{offspring}}{
        \tcc{\rmfamily\itshape\textcolor{gray}{replace first number with index}}
        offspring[index] $\leftarrow$ \FReSub{\str{\textbackslash d+}, str(index), line, 1}
    }

    \BlankLine
    \Return{\FJoin{offspring, \str{\textbackslash n}}} 
    \tcp{\rmfamily\itshape\textcolor{gray}{join lines into netlist string}}
}

\caption{Merge two netlists by line mixing.}
\label{alg:netlist_mixing}
\end{algorithm}

%% file: alg_component_merging.tex
\begin{algorithm}
\footnotesize
\DontPrintSemicolon
\SetAlgoNlRelativeSize{-1}       
\SetNlSty{}{}{:}    
\SetKwComment{Comment}{\textcolor{gray}{\#}$\,$}{}
\SetKwComment{tcc}{\textcolor{gray}{\#}$\,$}{}
\SetKwComment{tcp}{\textcolor{gray}{\#}$\,$}{}
\SetKwComment{docsign}{\textcolor{orange}{\texttt{"""}}}{}
\SetKwComment{doctext}{}{}
\SetKwFunction{FMergeComponents}{MixComponents}
\SetKwFunction{FReplace}{replace}
\SetKwFunction{FSplit}{split}
\SetKwFunction{FZip}{zip}
\SetKwFunction{FRandom}{RandomChoice}
\SetKwFunction{FAppend}{append}
\SetKwFunction{FJoin}{join}
\SetKwProg{Fn}{Function}{}{}

\Fn{\FMergeComponents{\newline component1: str, component2: str, force\_bulk: bool}}{
    \docsign{}
    \doctext{\rmfamily\itshape\textcolor{orange}{Merge two component definitions into new one.}}
    \doctext{}
    \doctext{\rmfamily\itshape\textcolor{orange}{We assume that `component1` and `component2` have same number of elements, i.e., length after splitting with `\textbackslash s*` as delimiter.}}
    \docsign{}

    \textbf{\textit{\textcolor{darkgreen}{Handle bulk connection to supply/ground (if forced by user)}}} \;
    \If{force\_bulk}{
        \tcc{\rmfamily\itshape\textcolor{gray}{join bulk connection and component name into one}}
        component1 $\leftarrow$ component1.\FReplace{\newline \str{net\_supply\_0 sky130\_fd\_pr\_\_pfet\_01v8}, \str{net\_supply\_0\#sky130\_fd\_pr\_\_pfet\_01v8}} \;
        component1 $\leftarrow$ component1.\FReplace{\newline \str{0 sky130\_fd\_pr\_\_nfet\_01v8}, \str{0\#sky130\_fd\_pr\_\_nfet\_01v8}} \;
        component2 $\leftarrow$ component2.\FReplace{\newline \str{net\_supply\_0 sky130\_fd\_pr\_\_pfet\_01v8}, \str{net\_supply\_0\#sky130\_fd\_pr\_\_pfet\_01v8}} \;
        component2 $\leftarrow$ component2.\FReplace{\newline \str{0 sky130\_fd\_pr\_\_nfet\_01v8}, \str{0\#sky130\_fd\_pr\_\_nfet\_01v8}} \;
    }

    \BlankLine
    \textbf{\textit{\textcolor{darkgreen}{Merge parts from both component lines}}} \;
    component $\leftarrow$ [ ] \Comment*[r]{\rmfamily\itshape\textcolor{gray}{initialize empty offspring}}
    component1 $\leftarrow$ component1.\FSplit{\str{ }} \Comment*[r]{\rmfamily\itshape\textcolor{gray}{split into elements}}
    component2 $\leftarrow$ component2.\FSplit{\str{ }}  \Comment*[r]{\rmfamily\itshape\textcolor{gray}{split into elements}}

    \BlankLine
    
    \ForEach{(e1, e2) in \FZip{component1 , component2}}{
        \tcc{\rmfamily\itshape\textcolor{gray}{randomly choose action for part (equal prob.)}}
        action $\leftarrow$ \FRandom{\str{first}, \str{second}} \;
        
        \BlankLine
        
        \If{action $==$ \str{first}}{
            component.\FAppend{e1} \tcp{\rmfamily\itshape\textcolor{gray}{Use component1 element}}
        }
        \Else{
            component.\FAppend{e2} \tcp{\rmfamily\itshape\textcolor{gray}{Use component2 element}}
        }
    }

    \BlankLine

    \tcc{\rmfamily\itshape\textcolor{gray}{join parts to generate component definition line}}
    component $\leftarrow$ \FJoin{component, \str{ }}     

    \BlankLine

    \textbf{\textit{\textcolor{darkgreen}{Handle bulk connection to supply/ground (if forced by user)}}} \;
    \If{force\_bulk}{
        \tcc{\rmfamily\itshape\textcolor{gray}{split into bulk and component}}
        component $\leftarrow$ component.\FReplace{\newline \str{net\_supply\_0\#sky130\_fd\_pr\_\_pfet\_01v8},\newline \str{net\_supply\_0 sky130\_fd\_pr\_\_pfet\_01v8}} \;
        component $\leftarrow$ component.\FReplace{\newline \str{0\#sky130\_fd\_pr\_\_nfet\_01v8},\newline \str{0 sky130\_fd\_pr\_\_nfet\_01v8}} \;
    }

    \BlankLine
    \Return{component}
}

\caption{Merge two components into one.}
\label{alg:component_mixing}
\end{algorithm}

%% file: alg_pruning.tex
\begin{algorithm}[t]
\footnotesize
\DontPrintSemicolon
\SetAlgoNlRelativeSize{-1}       
\SetNlSty{}{}{:}    
\SetKwComment{Comment}{\textcolor{gray}{\#}$\,$}{}
\SetKwComment{tcc}{\textcolor{gray}{\#}$\,$}{}
\SetKwComment{tcp}{\textcolor{gray}{\#}$\,$}{}
\SetKwComment{docsign}{\textcolor{orange}{\texttt{"""}}}{}
\SetKwComment{doctext}{}{}
\SetKwFunction{FRouletteWheelSelection}{roulette\_wheel\_selection}
\SetKwFunction{FSplit}{split}
\SetKwFunction{FDefaultDict}{defaultdict}
\SetKwFunction{FShuffle}{shuffle}
\SetKwFunction{FRandomPerm}{randElement}
\SetKwFunction{FMergeComponents}{MixComponents}
\SetKwFunction{FNormalizeNetlist}{NormalizeNetlist}
\SetKwFunction{Fenumerate}{enumerate}
\SetKwFunction{Fappend}{append}
\SetKwFunction{Fdel}{del}
\SetKwFunction{Fvalues}{values}
\SetKwFunction{Frange}{range}
\SetKwFunction{Fjoin}{join}
\SetKwFunction{Frandomnetlist}{SampleRandomNetlist}
\SetKwProg{Fn}{Function}{}{}

\Fn{PruneNetlist(\newline netlist: str, force\_bulk: bool, pruning\_ratio: float = 0.1)}{
    \docsign{}
    \doctext{\rmfamily\itshape\textcolor{orange}{Use pruning to generate a new offspring netlist.}}
    \docsign{}
    
    \textbf{\textit{\textcolor{darkgreen}{Initialization}}}
    
    \tcc{\rmfamily\itshape\textcolor{gray}{split netlist into lines}}
    lines $\leftarrow$ netlist.\FSplit{\str{\textbackslash n}} \;

    \tcc{\rmfamily\itshape\textcolor{gray}{determine number of pruning steps}}
    num\_pruning\_steps $\leftarrow$ 1 + int(pruning\_ratio $\cdot$ len(lines)) \;

    \BlankLine

    \textbf{\textit{\textcolor{darkgreen}{Perform pruning}}}
    
    \ForEach{step in \Frange{num\_pruning\_steps}}{
        \tcc{\rmfamily\itshape\textcolor{gray}{group lines by number of elements}}
        groups $\leftarrow$ \FDefaultDict{list} \;
        \ForEach{(i, line) in \Fenumerate{lines}}{
            groups[line.\textbf{count}(\str{ })].\Fappend{i} \;
        }
        \BlankLine

        \tcc{\rmfamily\itshape\textcolor{gray}{get list-of-lists with indices of equal length}}
        indices\_equal\_length $\leftarrow$ \newline [group for group in groups.\Fvalues{} \textbf{if} len(group) $>$ 1] \;

        \BlankLine

        \If{same\_length}{
            \tcc{\rmfamily\itshape\textcolor{gray}{randomly select two indices for pruning}}
            indices $\leftarrow$ \newline indices\_equal\_length.\FRandomPerm{}.\FShuffle{}[:2] \;

            \BlankLine
            \tcc{\rmfamily\itshape\textcolor{gray}{merge components and replace first line}}
            lines[idx[0]] $\leftarrow$ \FMergeComponents{\newline lines[indices[0]], lines[indices[1]], force\_bulk} \;

            \BlankLine
            \tcc{\rmfamily\itshape\textcolor{gray}{delete second line}}
            \Fdel lines[idx[1]] \;

            \BlankLine
            \tcc{\rmfamily\itshape\textcolor{gray}{normalize netlist and fix numbering}}
            netlist $\leftarrow$ \newline \FNormalizeNetlist{\Fjoin{lines, \str{\textbackslash n}}} \;
        }
        \Else{
            \tcc{\rmfamily\itshape\textcolor{gray}{Fallback to random if no lines that we can mix}}
            netlist $\leftarrow$ \str{ } \;
            \textbf{break} \;
        }
    }

    \BlankLine
    \Return{netlist} \tcp{\rmfamily\itshape\textcolor{gray}{Return pruned netlist}}
}

\caption{Pruning of lines in a netlist.}
\label{alg:pruning}
\end{algorithm}

%% file: alg_genetic_algorithm.tex
\begin{algorithm}
\footnotesize
\DontPrintSemicolon
\SetAlgoNlRelativeSize{-1}       
\SetNlSty{}{}{:}    
\SetKwComment{Comment}{\textcolor{gray}{\#}$\,$}{}
\SetKwComment{tcc}{\textcolor{gray}{\#}$\,$}{}
\SetKwComment{tcp}{\textcolor{gray}{\#}$\,$}{}
\SetKwFunction{FRandomChoice}{RandomChoice}
\SetKwFunction{FList}{list}
\SetKwFunction{FArgsort}{argsort}
\SetKwFunction{FArange}{arange}
\SetKwFunction{FRouletteSelection}{roulette\_wheel\_selection}
\SetKwFunction{FMergeNetlists}{MixNetlists}
\SetKwFunction{FNetlistToData}{netlist2data}
\SetKwFunction{FSampleRandom}{SampleRandomGraph}
\SetKwFunction{FDataToNetlist}{data2netlist}
\SetKwFunction{FToDevice}{to}
\SetKwFunction{FLinePruning}{LinePruning}
\SetKwComment{docsign}{\textcolor{orange}{\texttt{"""}}}{}
\SetKwComment{doctext}{}{}

\SetKwProg{Fn}{Function}{}{}
\Fn{RandomGraphSampler(hparams)}{
    \docsign{}
    \doctext{\rmfamily\itshape\textcolor{orange}{Sample circuit graph using uniform distributions \cite{uhlich2024graco}.}}
    \doctext{}
    \doctext{\rmfamily\itshape\textcolor{orange}{Optionally takes consistency checks into account which are set in `hparams`.}}
    \docsign{}}

\BlankLine

\Fn{GenerateNetlist(rewards: list, population: list, hparams)}{
    \docsign{}
    \doctext{\rmfamily\itshape\textcolor{orange}{Generate new offspring netlist using a genetic operation.}}
    \docsign{}

    \textbf{\textit{\textcolor{darkgreen}{Initialization}}}

    \tcc{\rmfamily\textcolor{gray}{determine population size}}
    N $\leftarrow$ len(population) 

    \tcc{\rmfamily\textcolor{gray}{randomly decide generation approach (equal probability)}}
    strategy $\leftarrow$ \newline \FRandomChoice{\str{crossover}, \str{mutation}, \str{pruning}}
    
    \BlankLine
    \tcc{\rmfamily\textcolor{gray}{get list of elite netlists (either $\eta$ or $\zeta$ is set)}}
    \If{$N\cdot\eta \geq 2$ \textbf{\emph{or}} $N \geq \zeta$}{
        \tcc{\rmfamily\textcolor{gray}{sort rewards (largest to smallest)}}
        sorted\_indices $\leftarrow$ \FArgsort{rewards, descending=True}

        \tcc{\rmfamily\textcolor{gray}{keep $N\cdot\eta$ best netlists}}
        elites $\leftarrow$ [netlists[i] \textbf{for} i \textbf{in} sorted\_indices[:$N // \eta$]]

        \tcc{\rmfamily\textcolor{gray}{generate rank vector for all elite netlists}}
        ranks $\leftarrow$ \FArange{start=N, end=0, step=-1}
    } \Else{
        \tcc{\rmfamily\textcolor{gray}{we do not have enough elite netlists yet}}
        elites $\leftarrow$ []
    }

    \BlankLine
    \textbf{\textit{\textcolor{darkgreen}{Generate new netlist based on chosen strategy}}} \;

    \If{elites \textbf{\emph{and}} strategy $==$ \str{crossover}}{
        \tcc{\rmfamily\textcolor{gray}{Select two parents using roulette-wheel sampling}}
        \tcc{\rmfamily\textcolor{gray}{(sampling with replacement, i.e., \emph{idx1 $==$ idx2} possible)}}
        idx1, idx2 $\leftarrow$ \linebreak \FRouletteSelection{ranks, n\_samples=2}

        parent1, parent2 $\leftarrow$ elites[idx1], elites[idx2]

        \BlankLine

        \tcc{\rmfamily\textcolor{gray}{Merge netlists}}
        offspring $\leftarrow$ \FMergeNetlists{parent1, parent2}

        \BlankLine

        \If{offspring}{
            \tcc{\rmfamily\textcolor{gray}{convert netlist to graph}}
            graph $\leftarrow$ \FNetlistToData{offspring}
        } \Else{
            \tcc{\rmfamily\textcolor{gray}{randomly generate graph as offspring was empty}}
            graph $\leftarrow$ \FSampleRandom{}
        }
    }
    \ElseIf{elites \textbf{\emph{and}} strategy $==$ \str{mutation}}{
        \tcc{\rmfamily\textcolor{gray}{select first parent from elite netlists}}
        idx $\leftarrow$ \linebreak \FRouletteSelection{ranks, n\_samples=1}

        parent1 $\leftarrow$ elites[idx]

        \tcc{\rmfamily\textcolor{gray}{randomly generate second parent}}
        parent2 $\leftarrow$ \FDataToNetlist{\FSampleRandom{}}

        \BlankLine

        \tcc{\rmfamily\textcolor{gray}{Merge netlists}}
        offspring $\leftarrow$ \FMergeNetlists{parent1, parent2}
    }
    \ElseIf{\textbf{\emph{not}} elites \textbf{\emph{or}} strategy $==$ \str{pruning}}{
        \tcc{\rmfamily\textcolor{gray}{select parent for pruning from elite netlists}}
        idx $\leftarrow$ \linebreak \FRouletteSelection{ranks, n\_samples=1}

        \tcc{\rmfamily\textcolor{gray}{prune lines}}
        offspring $\leftarrow$ \FLinePruning{elites[idx], force\_bulk}
    }
    
    \BlankLine

    \If{offspring}{
        \tcc{\rmfamily\textcolor{gray}{convert netlist to graph}}
        graph $\leftarrow$ \FNetlistToData{offspring}
    } \Else{
        \tcc{\rmfamily\textcolor{gray}{randomly generate graph as offspring was empty}}
        graph $\leftarrow$ \FSampleRandom{}
    }
        
    \BlankLine
    \Return{graph} \Comment*[r]{\rmfamily\textcolor{gray}{return generated graph (as PyG data)}}
}

\caption{Algorithmic description of SpiceMixer.}
\label{alg:netlist_generation}
\end{algorithm}

%% file: references.bib
@article{uhlich2024graco,
  title={{GraCo} -- {A} Graph Composer for Integrated Circuits},
  author={Uhlich, Stefan and Bonetti, Andrea and Venkitaraman, Arun and Momeni, Ali and Matsuo, Ryoga and Hsieh, Chia-Yu and Ohbuchi, Eisaku and Servadei, Lorenzo},
  journal={arXiv preprint arXiv:2411.13890},
  year={2024}
}

@article{matsuo2024schemato,
  title={Schemato -- {A}n LLM for Netlist-to-Schematic Conversion},
  author={Matsuo, Ryoga and Uhlich, Stefan and Venkitaraman, Arun and Bonetti, Andrea and Hsieh, Chia-Yu and Momeni, Ali and Mauch, Lukas and Capone, Augusto and Ohbuchi, Eisaku and Servadei, Lorenzo},
  journal={arXiv preprint arXiv:2411.13899},
  year={2024}
}

@article{momeni2025locality,
  title={Locality-aware Surrogates for Gradient-based Black-box Optimization},
  author={Momeni, Ali and Uhlich, Stefan and Venkitaraman, Arun and Hsieh, Chia-Yu and Bonetti, Andrea and Matsuo, Ryoga and Ohbuchi, Eisaku and Servadei, Lorenzo},
  journal={arXiv preprint arXiv:2501.19161},
  year={2025}
}

@inproceedings{kool2019buy,
  title={Buy 4 reinforce samples, get a baseline for free!},
  author={Kool, Wouter and van Hoof, Herke and Welling, Max},
  booktitle={Proceedings of the ICLR 2019 Workshop: Deep RL Meets Structured Prediction},
  year={2019}
}

@article{williams1992simple,
  title={Simple statistical gradient-following algorithms for connectionist reinforcement learning},
  author={Williams, Ronald J},
  journal={Machine learning},
  volume={8},
  pages={229--256},
  year={1992},
  publisher={Springer}
}

@article{salimans2017evolution,
  title={Evolution strategies as a scalable alternative to reinforcement learning},
  author={Salimans, Tim and Ho, Jonathan and Chen, Xi and Sidor, Szymon and Sutskever, Ilya},
  journal={arXiv preprint arXiv:1703.03864},
  year={2017}
}

@article{campilho2024analog,
  title={Analog flat-level circuit synthesis with genetic algorithms},
  author={Campilho-Gomes, Miguel and Tavares, Rui and Goes, Jo{\~a}o},
  journal={IEEE Access},
  year={2024},
  publisher={IEEE}
}

@article{hansen2016cma,
  title={The {CMA} evolution strategy: A tutorial},
  author={Hansen, Nikolaus},
  journal={arXiv preprint arXiv:1604.00772},
  year={2016}
}

@inproceedings{lai2025analogcoder,
  title={{AnalogCoder}: Analog circuit design via training-free code generation},
  author={Lai, Yao and Lee, Sungyoung and Chen, Guojin and Poddar, Souradip and Hu, Mengkang and Pan, David Z and Luo, Ping},
  booktitle={Proceedings of the AAAI Conference on Artificial Intelligence},
  volume={39},
  number={1},
  pages={379--387},
  year={2025}
}

@article{gao2025analoggenie,
  title={{AnalogGenie}: A Generative Engine for Automatic Discovery of Analog Circuit Topologies},
  author={Gao, Jian and Cao, Weidong and Yang, Junyi and Zhang, Xuan},
  journal={arXiv preprint arXiv:2503.00205},
  year={2025}
}

@book{razavi2017design,
  author = {Razavi, Behzad},
  edition = {2nd},
  publisher = {McGraw Hill},
  title = {Design of analog CMOS integrated circuit},
  year = 2017
}

@book{jespers2017systematic,
  title={Systematic Design of Analog {CMOS} Circuits},
  author={Jespers, P.G.A. and Murmann, B.},
  year={2017},
  publisher={Cambridge University Press}
}

@inproceedings{lyu2024study,
  title={A study on exploring and exploiting the high-dimensional design space for analog circuit design automation},
  author={Lyu, Ruiyu and Meng, Yuan and Zhao, Aidong and Bi, Zhaori and Zhu, Keren and Yang, Fan and Yan, Changhao and Zhou, Dian and Zeng, Xuan},
  booktitle={2024 29th Asia and South Pacific Design Automation Conference (ASP-DAC)},
  pages={671--678},
  year={2024},
  organization={IEEE}
}

@article{yang2017smart,
  title={{Smart-MSP}: A self-adaptive multiple starting point optimization approach for analog circuit synthesis},
  author={Yang, Yishi and Zhu, Hengliang and Bi, Zhaori and Yan, Changhao and Zhou, Dian and Su, Yangfeng and Zeng, Xuan},
  journal={IEEE Transactions on Computer-Aided Design of Integrated Circuits and Systems},
  volume={37},
  number={3},
  pages={531--544},
  year={2017},
  publisher={IEEE}
}

@inproceedings{settaluri2020autockt,
  title={{AutoCkt}: Deep reinforcement learning of analog circuit designs},
  author={Settaluri, Keertana and Haj-Ali, Ameer and Huang, Qijing and Hakhamaneshi, Kourosh and Nikolic, Borivoje},
  booktitle={2020 Design, Automation \& Test in Europe Conference \& Exhibition (DATE)},
  pages={490--495},
  year={2020},
  organization={IEEE}
}

@inproceedings{budak2021dnn,
  title={{DNN-Opt}: An rl inspired optimization for analog circuit sizing using deep neural networks},
  author={Budak, Ahmet F and Bhansali, Prateek and Liu, Bo and Sun, Nan and Pan, David Z and Kashyap, Chandramouli V},
  booktitle={2021 58th ACM/IEEE Design Automation Conference (DAC)},
  pages={1219--1224},
  year={2021},
  organization={IEEE}
}

@inproceedings{budak2023apostle,
  title={{APOSTLE}: Asynchronously parallel optimization for sizing analog transistors using DNN learning},
  author={Budak, Ahmet F and Smart, David and Swahn, Brian and Pan, David Z},
  booktitle={Proceedings of the 28th Asia and South Pacific Design Automation Conference},
  pages={70--75},
  year={2023}
}

@article{dong2023cktgnn,
  title={{CktGNN}: Circuit graph neural network for electronic design automation},
  author={Dong, Zehao and Cao, Weidong and Zhang, Muhan and Tao, Dacheng and Chen, Yixin and Zhang, Xuan},
  journal={arXiv preprint arXiv:2308.16406},
  year={2023}
}

@article{meissner2014feats,
  title={{FEATS}: Framework for explorative analog topology synthesis},
  author={Meissner, Markus and Hedrich, Lars},
  journal={IEEE Transactions on Computer-Aided Design of Integrated Circuits and Systems},
  volume={34},
  number={2},
  pages={213--226},
  year={2014},
  publisher={IEEE}
}

@article{zhao2020automated,
  title={An automated topology synthesis framework for analog integrated circuits},
  author={Zhao, Zhenxin and Zhang, Lihong},
  journal={IEEE Transactions on Computer-Aided Design of Integrated Circuits and Systems},
  volume={39},
  number={12},
  pages={4325--4337},
  year={2020},
  publisher={IEEE}
}

@article{zadeh2025generative,
  title={Generative AI for Analog Integrated Circuit Design: Methodologies and Applications},
  author={Zadeh, Danial Noori and Elamien, Mohamed B},
  journal={IEEE Access},
  year={2025},
  publisher={IEEE}
}

@misc{hansen2019pycma,
  author       = {Nikolaus Hansen and Youhei Akimoto and Petr Baudis},
  title        = {{CMA-ES/pycma} on {G}ithub},
  howpublished = {Zenodo, DOI:10.5281/zenodo.2559634},
  month        = feb,
  year         = 2019,
  doi          = {10.5281/zenodo.2559634},
  url          = {https://doi.org/10.5281/zenodo.2559634},
}

@phdthesis{trefzer2006evolution,
  title={Evolution of transistor circuits},
  author={Trefzer, Martin Albrecht},
  year={2006}
}

@inproceedings{kruiskamp1995darwin,
  title={DARWIN: CMOS opamp synthesis by means of a genetic algorithm},
  author={Kruiskamp, Wim and Leenaerts, Domine},
  booktitle={Proceedings of the 32nd annual ACM/IEEE design automation conference},
  pages={433--438},
  year={1995}
}

@article{kwon2023circuit,
  title={Circuit-centric Genetic Algorithm (CGA) for Analog and Radio-Frequency Circuit Optimization},
  author={Kwon, Mingi and Lee, Yeonjun and Song, Ickhyun},
  journal={arXiv preprint arXiv:2403.17938},
  year={2023}
}

@article{rashid2024machine,
  title={Machine learning driven global optimisation framework for analog circuit design},
  author={Rashid, Ria and Krishna, Komala and George, Clint Pazhayidam and Nambath, Nandakumar},
  journal={Microelectronics Journal},
  volume={151},
  pages={106362},
  year={2024},
  publisher={Elsevier}
}

@article{yengui2012hybrid,
  title={A hybrid GA-SQP algorithm for analog circuits sizing},
  author={Yengui, Firas and Labrak, Lioua and Frantz, Felipe and Daviot, Renaud and Abouchi, Nacer and O’Connor, Ian},
  journal={Circuits and Systems},
  volume={3},
  number={2},
  pages={146--152},
  year={2012},
  publisher={Scientific Research Publishing}
}

@article{barari2014analog,
  title={Analog circuit design optimization based on evolutionary algorithms},
  author={Barari, Mansour and Karimi, Hamid Reza and Razaghian, Farhad},
  journal={Mathematical Problems in Engineering},
  volume={2014},
  number={1},
  pages={593684},
  year={2014},
  publisher={Wiley Online Library}
}

@article{torres2010robust,
  title={A Robust Evolvable System for the Synthesis of Analog Circuits.},
  author={Torres Soto, Aurora and Ponce de Le{\'o}n Sent{\'\i}, Eunice E and Hern{\'a}ndez Aguirre, Arturo and Torres Soto, Mar{\'\i}a Dolores and D{\'\i}az D{\'\i}az, Elva},
  journal={Computaci{\'o}n y Sistemas},
  volume={13},
  number={4},
  pages={409--421},
  year={2010}
}

@article{sapargaliyev2010challenging,
  title={Challenging the evolutionary strategy for synthesis of analogue computational circuits},
  author={Sapargaliyev, Yerbol A and Kalganova, Tatiana G},
  year={2010},
  publisher={Scientific Research Publishing}
}

@inproceedings{das2007gapsys,
  title={GAPSYS: A GA-based tool for automated passive analog circuit synthesis},
  author={Das, Angan and Vemuri, Ranga},
  booktitle={2007 IEEE International Symposium on Circuits and Systems (ISCAS)},
  pages={2702--2705},
  year={2007},
  organization={IEEE}
}

@inproceedings{noren2001analog,
  title={Analog circuit design using genetic algorithms},
  author={Noren, Kenneth V and Ross, John E},
  booktitle={Second Online Symposium for Electronics Engineers},
  year={2001}
}

@inproceedings{ning1991seas,
  title={SEAS: A simulated evolution approach for analog circuit synthesis},
  author={Ning, Zhen-Qiu and Mouthaan, AJ and Wallinga, Hans},
  booktitle={IEEE Custom Integrated Circuits Conference, CICC'91, San Diego, USA, 12-15 May, 1991: Proceedings IEEE Custom Integrated Circuits Conference (CICC'91)},
  pages={5--2},
  year={1991},
  organization={IEEE}
}

@incollection{koza1996automated,
  title={Automated design of both the topology and sizing of analog electrical circuits using genetic programming},
  author={Koza, John R and Bennett III, Forrest H and Andre, David and Keane, Martin A},
  booktitle={Artificial intelligence in design’96},
  pages={151--170},
  year={1996},
  publisher={Springer}
}

@article{zhang2024analogxpert,
  title={AnalogXpert: Automating Analog Topology Synthesis by Incorporating Circuit Design Expertise into Large Language Models},
  author={Zhang, Haoyi and Sun, Shizhao and Lin, Yibo and Wang, Runsheng and Bian, Jiang},
  journal={arXiv preprint arXiv:2412.19824},
  year={2024}
}

@inproceedings{zhao2022deep,
  title={Deep reinforcement learning for analog circuit structure synthesis},
  author={Zhao, Zhenxin and Zhang, Lihong},
  booktitle={2022 Design, Automation \& Test in Europe Conference \& Exhibition (DATE)},
  pages={1157--1160},
  year={2022},
  organization={IEEE}
}

@misc{skywater130pdk,
  author       = {{Google and SkyWater Technology Foundry}},
  title        = {SkyWater 130nm {PDK}},
  year         = {2020},
  url          = {https://github.com/google/skywater-pdk}
}

@misc{alberts14,
  title={Molecular Biology of the Cell},
  author={Alberts, B. and Johnson, A. and Lewis, J. and Morgan, D. and Raff, M. and Roberts, K. and Walter, P.},
  year={2014},
  publisher={Garland Science}
}

@article{shi2021sizing,
  title={Sizing of multi-stage Op Amps by combining design equations with the gm/ID method},
  author={Shi, Guoyong},
  journal={Integration},
  volume={79},
  pages={48--60},
  year={2021},
  publisher={Elsevier}
}
